\documentclass[]{siamart}

\usepackage{lipsum}
\usepackage{amsfonts}
\usepackage{graphicx}
\usepackage{epstopdf}
\usepackage{algorithm}
\usepackage{algpseudocode}
\usepackage{pgfplots}
\pgfplotsset{compat=1.13}
\usepackage{tikz}
\usepackage{todonotes}
\usepackage{booktabs}
\usepackage{multirow} 
\usepackage{subfigure}

\usepackage{etoolbox}
\usepackage[binary-units=true]{siunitx}
\robustify\bfseries
\usepackage{amsopn}

\makeatletter
\def\blfootnote{\gdef\@thefnmark{}\@footnotetext}
\makeatother

\newsiamremark{remark}{Remark}

\ifpdf
  \DeclareGraphicsExtensions{.eps,.pdf,.png,.jpg}
\else
  \DeclareGraphicsExtensions{.eps}
\fi

\newcommand{\TheTitle}{A matrix-free approach to parallel and memory-efficient deformable image registration} 

\newcommand{\TheAuthors}{L. K\"onig, J. R\"uhaak, A. Derksen and J. Lellmann}

\headers{Matrix-free deformable image registration}{\TheAuthors}

\title{{\TheTitle}}

\author{
  Lars K\"onig\thanks{Fraunhofer MEVIS, L\"ubeck, Germany}
    (\email{lars.koenig@mevis.fraunhofer.de}, \email{jan.ruehaak@mevis.fraunhofer.de}, \email{alexander.derksen@mevis.fraunhofer.de}).
  \and
  Jan R\"uhaak\footnotemark[1]
  \and
  Alexander Derksen\footnotemark[1]
  \and
  Jan Lellmann$^{*,}$\thanks{Institute of Mathematics and Image Computing, University of L\"ubeck, L\"ubeck, Germany}
  (\email{jan.lellmann@mic.uni-luebeck.de})
}

\newcommand{\iMinusX}{i-x}
\newcommand{\iMinusY}{i-y}
\newcommand{\iMinusZ}{i-z}
\newcommand{\iPlusX} {i+x}
\newcommand{\iPlusY} {i+y}
\newcommand{\iPlusZ} {i+z}

\newcommand{\iMinusK}{i-k}
\newcommand{\iPlusK} {i+k}

\newcommand{\iY}{i}
\newcommand{\jY}{j}
\newcommand{\kY}{k}

\newcommand{\trafo}{\varphi}
\newcommand{\x}{\mathrm{x}}
\newcommand{\y}{\mathrm{y}}

\newcommand{\rr}{\mathbb{R}}
\newcommand{\R}{\mathcal{R}}
\newcommand{\T}{\mathcal{T}}
\newcommand{\omr}{\Omega_{\R}}
\newcommand{\vecx}{\mathbf{x}}
\newcommand{\vecy}{\mathbf{y}}
\newcommand{\vecu}{\mathbf{u}}
\newcommand{\barmy}{\bar m^{\mathrm y}}
\newcommand{\my}{m^{\mathrm y}}
\newcommand{\hy}{h^{\mathrm y}}
\newcommand{\barhy}{\bar h^{\mathrm y}}

\newcommand{\vecxy}{\vecx^{\mathrm y}}

\providecolor{redlk}{RGB}{255,65,14}
\providecolor{FhCD02}{RGB}{136,188,226}
\providecolor{FhCD19}{RGB}{255,220,0}

\ifpdf
\hypersetup{
  pdftitle={\TheTitle},
  pdfauthor={\TheAuthors}
}
\fi

\begin{document}

\maketitle

\begin{abstract}
We present a novel computational approach to fast and memory-efficient deformable image registration. In the variational registration model, the computation of the objective function derivatives is the computationally most expensive operation, both in terms of runtime and memory requirements.
In order to target this bottleneck, we 
analyze the matrix structure of gradient and Hessian computations for the case of the normalized gradient fields distance measure and curvature regularization. 
Based on this analysis, we derive equivalent matrix-free closed-form expressions for derivative computations, eliminating the need for storing intermediate results and the costs of sparse matrix arithmetic. This has further benefits: (1) matrix computations can be fully parallelized,  (2) memory complexity for derivative computation is reduced from linear to constant, and (3) overall computation times are substantially reduced.

In comparison with an optimized matrix-based reference implementation, the CPU implementation achieves speedup factors between 3.1 and 9.7, and we are able to handle substantially higher resolutions. Using a GPU implementation, we achieve an additional speedup factor of up to 9.2.

Furthermore, we evaluated the approach on real-world medical datasets. On ten publicly available lung CT images from the DIR-Lab 4DCT dataset, we achieve the best mean landmark error of \SI{0.93}{\milli\meter} compared to other submissions on the DIR-Lab website, with an average runtime of only~\SI{9.23}{\second}. 
Complete non-rigid registration of full-size 3D thorax-abdomen CT volumes from oncological follow-up 
is achieved in \SI{12.6}{\second}. The experimental results show that the proposed matrix-free algorithm enables the use of variational registration models also in applications which were previously impractical due to memory or runtime restrictions.
\end{abstract}



\section{Introduction}
\blfootnote{$\!\!\!\!\!^\ddag$First preliminary results of this work were announced in 4-page conference proceedings \cite{konig2014fast,konig2015parallel}}
Image registration denotes the process of aligning two or more images for analysis and comparison~\cite{lemoigne2011image}. \emph{Deformable} registration approaches, which allow for non-rigid, non-linear deformations, are of particular interest in many areas of medical imaging and contribute to the development of new technologies for diagnosis and therapy. Applications range from motion correction in gated cardiac Positron Emission Tomography (PET)~\cite{gigengack2012motion} and biomarker computation in regional lung function analysis~\cite{galban2012computed} to inter-subject registration for automated labeling of brain data~\cite{avants2008symmetric}. 

\begin{figure}[t]
\centering
\subfigure[Prior scan]{\includegraphics[width=0.24\textwidth]{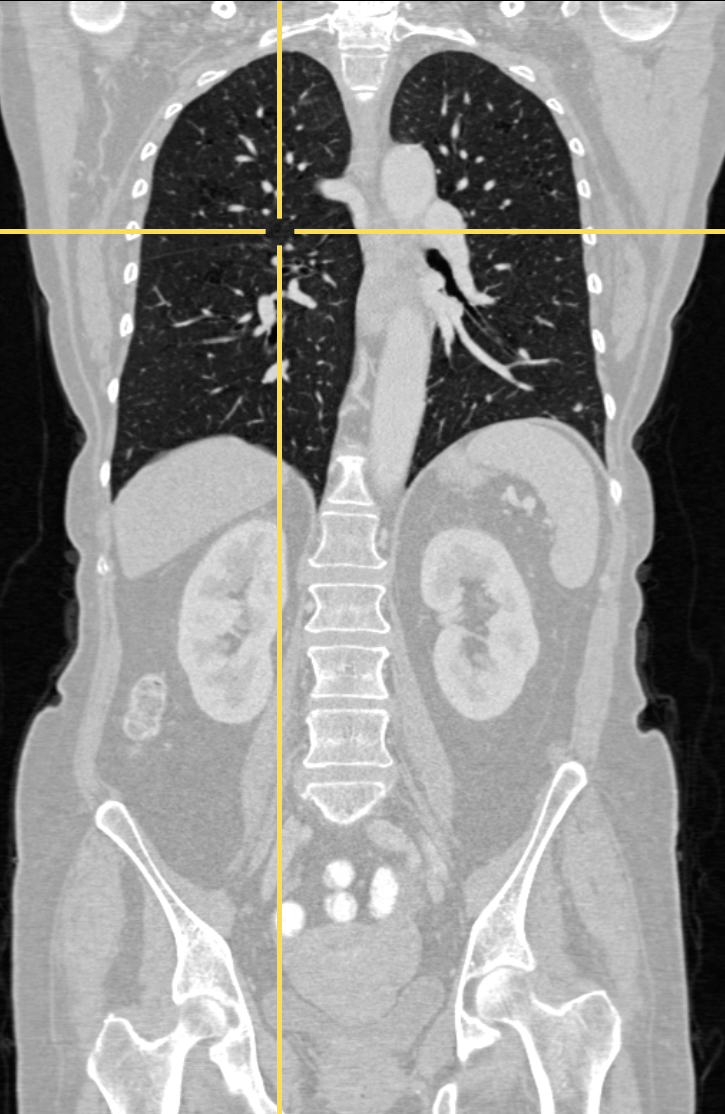}}
\subfigure[Current scan]{\includegraphics[width=0.24\textwidth]{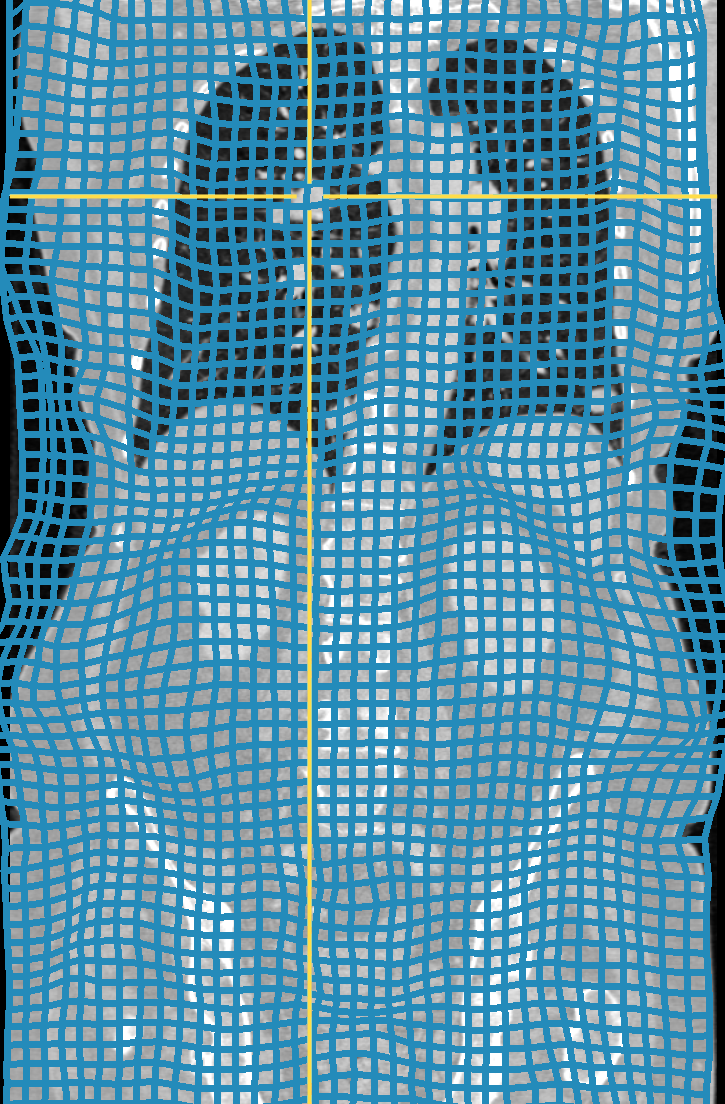}}
\subfigure[Initial difference]{\includegraphics[width=0.24\textwidth]{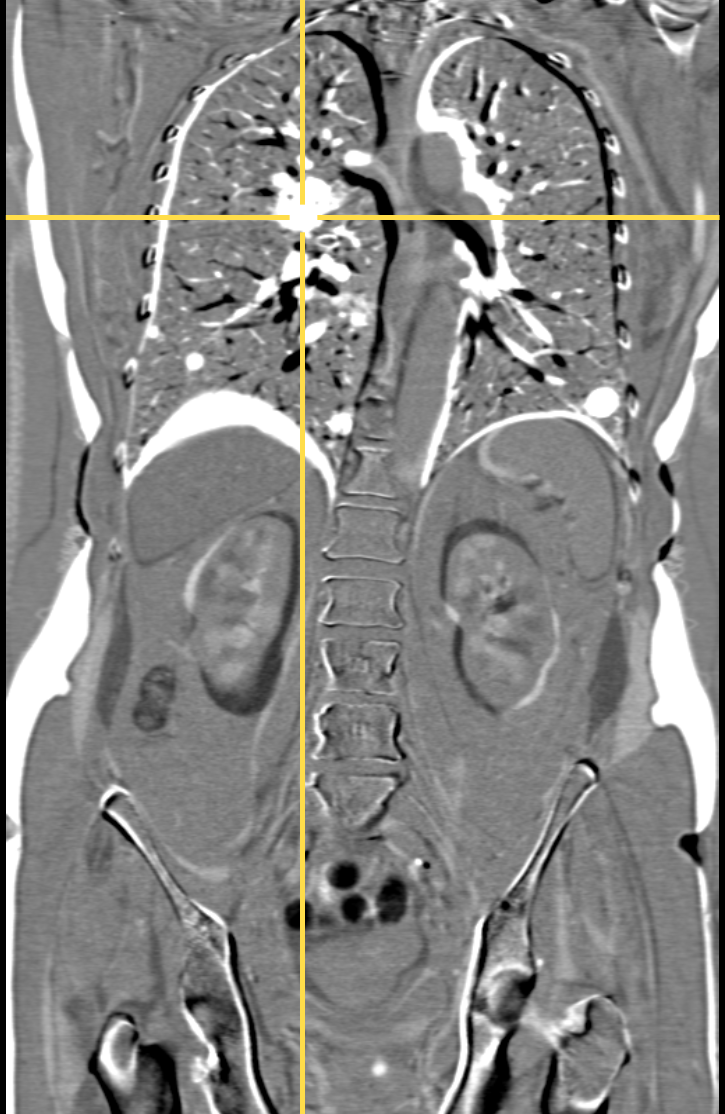}}
\subfigure[After registration]{\includegraphics[width=0.24\textwidth]{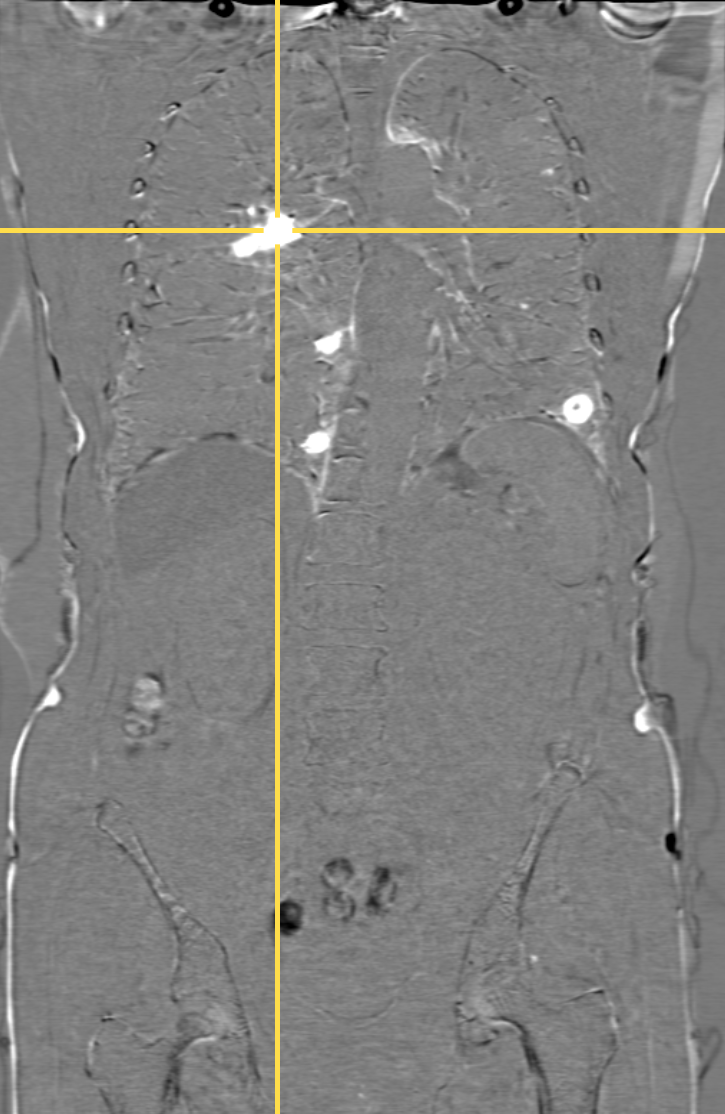}}
\caption{Registration of thorax-abdomen 3D CT scans for oncological follow-up. (a) coronal slice of prior scan, (b) coronal slice of current scan with deformation grid overlay, (c) subtraction image before registration, (d) subtraction image after registration. The current scan is deformed onto the prior scan using the proposed variational method. The subtraction image highlights the areas of change (white spots in the lung) corresponding to tumor growth. Image courtesy of Radboud University Medical Center, Nijmegen, The Netherlands.}\label{fig:thoraxReg}
\end{figure}

For successful clinical adoption of deformable image registration, moderate memory consumption and low runtimes are indispensable, which is made more difficult by the fact that data is usually three-dimensional and therefore large. For example, in order to fuse computed tomography (CT) images for oncological follow-up,
volumes with typical sizes of around $512\times512\times900$ voxels need to undergo full 3D registration; see Figure~\ref{fig:thoraxReg} for an example of a thorax-abdomen registration. In a clinical setting, these registrations have to be performed for every acquired volume and must be available quickly in order not to slow down the clinical workflow.
In other situations, such as in population studies or during screening, the large number of required registrations 
 demands efficient deformable image registration algorithms. In some areas, such as liver ultrasound tracking, real-time performance is even required~\cite{konig2014nonlinear,deluca2015liver}.

Consequently, a lot of research has been dedicated to increasing the efficiency of image registration algorithms, see \cite{shackleford2013high,shams2010survey, eklund2013medical} for an overview. The literature can be divided into approaches that aim at modifying the algorithmic structure of a registration method in order to increase efficiency~\cite{haber2007octree,haber2008adaptive,sturmer2008fast} and approaches that specialize a given algorithm on a particular hardware platform -- mostly on massively parallel architectures such as Graphics Processing Units (GPUs)~\cite{kohn2006gpu,muyan2008fast,bui2009performance,shamonin2013fast}, but also on Digital Signal Processors (DSPs)~\cite{berg2014highly} or Field Programmable Gate Arrays (FPGAs)~\cite{castro2003fair}. 

\noindent\emph{Contributions and overview.}
In the following, we focus on variational approaches for image registration, which rely on numerical minimization of an energy function that depends on the input data. While such methods have been successfully used in various applications \cite{ruhaak2013highly,burger2013hyperelastic,gigengack2012motion,FAIR}, they involve expensive derivative computations, which are not straightforward to implement efficiently and hinder parallelizability.

In this work, we aim to work towards closing this gap:
\begin{itemize}
  \item We analyze the derivative structure 
  of the classical, widely-known variational image registration model~\cite{modersitzki2004, FAIR} with the normalized gradient fields distance measure~\cite{haber2005beyond} and curvature regularization~\cite{fischer2003curvature}, which has been successfully used for various image registration problems~\cite{stutzer2016evaluation,konig2016deformable,deluca2015liver} (Section \ref{sec:DeriComp} and Section \ref{sec:MatrixFree}). 
  \item We study the use of efficient \emph{matrix-free techniques} for derivative calculations in order to improve computational efficiency (Section \ref{sec:MatrixFree}). In particular, we present fully matrix-free computation rules, based on the work on affine-linear image registration in~\cite{ruhaak2017matrix} and deformable registration in~\cite{konig2014fast,konig2015parallel}, for objective function gradient computations (Section \ref{sec:mf_methods_ngf_gradient}) and Gauss-Newton Hessian-vector multiplications (Section \ref{sec:hessian-vector_multiplication}).
 \item We perform a theoretical analysis  of the proposed approach in terms of computational effort and memory usage (Section \ref{sec:algorithm_analysis}).
 \item Finally, we quantitatively evaluate CPU and GPU implementations (Section~\ref{sec:gpu_implementation}) of the new method with respect to speedup, parallel scalability (Section \ref{sec:ex_scalability}) and problem size, and in comparison with two alternative methods (Section \ref{sec:runtime_comparison}).
We demonstrate the real-world applicability on two medical applications (Section \ref{sec:medical_applications}).
\end{itemize}
Benefits of the proposed approach are
full parallelization of objective function and derivative computations,
reduction of derivative computation memory requirements from linear to constant,
as well as a large reduction in overall runtime.
The strategy can also be applied to other distance measures and regularizers, for which we discuss requirements and limitations. 

In order to underline the merits of the approach in the clinic, we exemplarily consider two medical applications. Firstly, the registration algorithm is employed for registering lung CT images in inhaled and exhaled position (Section \ref{sec:pulmonary_image_registration}).
On the widely-used DIR-Lab 4DCT database~\cite{castillo2009framework, castillo2010four}, we achieve the best mean landmark error of \SI{0.93}{\milli\meter} in comparison with the results reported at \cite{dirlab} with an average runtime of only \SI{9.23}{\second}, computed on a standard workstation. Secondly, we consider oncological follow-up studies in the thorax-abdomen region (Section \ref{sec:oncological_followup}), where qualitatively convincing results are obtained at a clinically feasible runtime of \SI{12.6}{\second}. 

\section{Variational image registration framework}\label{sec:VarImReg}

\subsection{Continuous model}
While there are numerous different applications of image registration, the goal is always the same: to obtain spatial correspondences between two or more images. Here we only consider the case of two images. The first image is denoted the \emph{reference image} $\mathcal R$ (sometimes also called fixed image), the second image is the \emph{template image} $\mathcal T$ (or moving image)~\cite{FAIR}. For three-dimensional gray-scale images, the two images are given by functions $\mathcal R: \mathbb R^3 \to \mathbb R$ and $\mathcal T: \mathbb R^3 \to \mathbb R$ on domains $\Omega_\mathcal R \subset \mathbb R^3$ and $\Omega_\mathcal T \subset \mathbb R^3$, mapping each coordinate to a gray value. To establish correspondence between the images, a \emph{deformation} function $\varphi:\Omega_\mathcal R \to \mathbb R^3$ is sought, which encodes the spatial alignment by mapping points from reference to template image domain. With the composition $\mathcal T(\varphi): \Omega_\mathcal R \to \mathbb R$, $x \mapsto \mathcal T(\varphi(x))$,
a \emph{deformed template} image in the reference image domain can be obtained.

In order to find a suitable deformation $\varphi$, two competing criteria must be balanced. Firstly, the deformed template image should be similar to the reference image, determined by some \emph{distance measure} $\mathcal D(\mathcal R, \mathcal T(\varphi))$. Secondly, as the distance measure alone is generally not enough to make the problem well-posed and robust, a \emph{regularizer} $\mathcal S(\varphi)$ is introduced, which requires the deformation to be ``reasonable''. An optimal deformation meeting both criteria is then found by solving the optimization problem 
\begin{align}
\min_{\varphi:\Omega_{\mathcal R} \to \mathbb{R}^3} \mathcal J(\varphi), \quad  \mathcal J(\varphi) := \mathcal D(\mathcal R, \mathcal T (\varphi)) + \alpha \mathcal S(\varphi), \label{eq:optimizationProblem}
\end{align}
where $\alpha > 0$ is a weighting parameter.
Numerous different approaches and definitions for distance measures~\cite{collignon1995automated,viola1997alignment,modersitzki2004} and regularizers~\cite{fischer2003curvature,broit1981optimal,burger2013hyperelastic} have been presented, each with specific properties tailored to the application.

This work focuses on the normalized gradient fields (NGF) distance measure. It was first presented in~\cite{haber2005beyond} and has since been successfully used in a wide range of applications~\cite{ruhaak2013highly,polzin2016memory,konig2016deformable}. The NGF distance is particularly suited for multi-modal registration problems, where image intensities of reference and template image are related in complex and usually non-linear ways, 
and represents a fast and robust alternative \cite{haber2005beyond} to the widely-used mutual information~\cite{collignon1995automated,viola1997alignment}. It is defined as
\begin{align}
\mathcal D(\varphi) = \int_{\Omega_{\mathcal R}}  1 - \left( \frac{\langle \nabla \mathcal T(\varphi),\nabla \mathcal R \rangle + \tau \varrho}{\Vert \nabla \mathcal T(\varphi) \Vert_\tau \Vert \nabla \mathcal R \Vert_\varrho}\right)^2 \mathrm dx,\label{eq:ngf}
\end{align}
where $\Vert \cdot \Vert_\varepsilon := \sqrt{\langle \cdot,\cdot \rangle + \varepsilon^2}$, and $\tau,\varrho > 0$ are parameters that control filtering of noise in each image. The underlying assumption is that even in different imaging modalities, intensity \emph{changes} still take place at corresponding locations. Therefore, the NGF distance advocates parallel alignment of \emph{image gradient directions}, i.e., edge orientations.

For the regularizer $\mathcal S$, we use \emph{curvature regularization}~\cite{fischer2003curvature},
\begin{align*}
 \mathcal S(\varphi) = \int_{\Omega_R} \sum_{j=1}^3 (\Delta u_j)^2 \mathrm dx.
\end{align*}
$\mathcal S$ is defined in terms of the \emph{displacement} $u := (u_1,u_2,u_3)^\top := \varphi - \mathrm{id}$, where id is the identity transformation. The curvature regularizer penalizes second-order derivatives and thus favors ``smooth'' deformations, while not penalizing affine transformations such as translations or rotations.
Similar to NGF, curvature regularization has been successfully used in multiple applications, e.g.,~\cite{ruhaak2013highly,konig2016deformable,konig2014nonlinear}.

\subsection{Discretization} \label{sec:discretization}

In order to solve the registration problem \eqref{eq:optimizationProblem}, we follow a \emph{discretize-then-optimize} approach~\cite{FAIR}: Firstly, the deformation $\varphi$ and objective function $\mathcal J$ are turned into a discretized energy. Secondly, this discretized energy is minimized using standard methods from numerical optimization (Section \ref{sec:numerical-optimization}). For 
reference, the notation introduced in the following is summarized in Table~\ref{table:notation}.

The \emph{image domain} $\Omega_\mathcal R$ is discretized on a cuboid grid with $m_x, m_y, m_z$ grid cells (voxels) in each dimension and $\bar m := m_x m_y m_z$ cells in total. The corresponding grid spacings are denoted by $h_x,h_y,h_z$, and the cell volume by $\bar h := h_x h_y h_z$.
The associated  \emph{image grid} is defined as the \emph{cell-centers}
\begin{align*}
\mathcal X := \left\{\left(\hat ih_x - \textstyle\frac{h_x}{2}, \hat jh_y - \frac{h_y}{2}, \hat kh_z - \frac{h_z}{2}\right)\!\,\Big\vert\, \hat i = 1,\ldots, m_x, \hat j = 1,\ldots, m_y, \hat k = 1,\ldots, m_z \right\}\!.
\end{align*}
Using a lexicographic ordering defined by the index mapping $i := \hat i +\hat j\, m_x + \hat k\, m_x m_y$, the elements in $\mathcal X$ can be assembled into a single vector $\mathbf{x} := ( \mathrm x_1,\ldots, \mathrm x_{3\bar m}) \in \mathbb R^{3\bar m}$, which is obtained by first storing all $x$-, then all $y$- and finally all $z$- coordinates in the order prescribed by $i$. 
The $i$-th grid point in $\mathcal X$ is then $\mathbf{x}_{i} := \left( \mathrm x_i, \mathrm x_{i+\bar m}, \mathrm x_{i+2\bar m}\right)\in \mathbb R^{3}$. Evaluating the reference image on all grid points can be compactly written as $R(\mathbf{x}) := \mathcal R\left(\mathbf{x}_{i}\right)_{i=1,\ldots,\bar m}$, resulting in a vector $R(\mathbf{x}) \in \mathbb R^{\bar m}$ of image intensities at the grid points.

In a similar way, we define the discretized \emph{deformation} $\mathbf y := \varphi(\mathbf x^\mathrm y)$ as the evaluation of $\varphi$ on a \emph{deformation grid} $\textbf x^\mathrm y$. Again we order all components of $\mathbf y$ in a vector, $\mathbf y = (\mathrm y_1, \ldots, \mathrm y_{3\bar m^\mathrm y})$, so that $\varphi(\mathbf x^\mathrm y_i) = ( \mathrm y_{\iY}, \mathrm y_{\iY+\bar m^\mathrm y}, \mathrm y_{\iY+2\bar m^\mathrm y})$, where the deformation grid has grid spacings $h^{\mathrm y}_x,h^{\mathrm y}_y,h^{\mathrm y}_z$, and $m_x^\mathrm y,m_y^ \mathrm y,m_z^ \mathrm y$ denote the number of grid points in each direction with $\bar m^\mathrm y := m_x^\mathrm y m_y^\mathrm y m_z^\mathrm y$. 

Importantly, instead of using cell-center coordinates, the deformation is discretized on cell corners, i.e., using a \emph{nodal} grid. This ensures that the deformation is discretized up to the boundary of the image domain, 
enabling an efficient grid conversion without requiring extrapolation or additional boundary conditions (Section \ref{sec:grid-conversion}).

This results in $m_x^\mathrm y,m_y^ \mathrm y,m_z^ \mathrm y$ coordinates in each direction and $\bar m^\mathrm y := m_x^\mathrm y m_y^\mathrm y m_z^\mathrm y$ grid points in total. The set of all nodal grid points is given by 
\begin{align*}
\mathcal X^\mathrm y\! := \Big\{\!\left(\hat \iY h^{\mathrm y}_x, \hat \jY h^{\mathrm y}_y,\hat \kY h^{\mathrm y}_z\right)\Big\vert\ \hat \iY\! = 0,\ldots, m^{\mathrm y}_x\!-\!1, \hat \jY\! = 0,\ldots, m^{\mathrm y}_y\!-\!1,\hat \kY\! = 0,\ldots, m^{\mathrm y}_z\!-\!1\Big\}. 
\end{align*}
Analogously to the image grid, using the linear index $\iY := \hat \iY +\hat \jY\, m_x^\mathrm y + \hat \kY\, m_x^\mathrm y m_y^\mathrm y$, the points from $\mathcal X^\mathrm y$ can be ordered lexicographically in a  
vector $\mathbf x^\mathrm y:= (\mathrm x^\mathrm y_1,\ldots,\mathrm x^\mathrm y_{3\bar m^{\mathrm y}}) \in \mathbb R^{3 \bar m^\mathrm y}$, 
with a single point having the coordinates $\mathbf x^\mathrm y_{\iY} := (\mathrm x^\mathrm y_{\iY}, \mathrm x^\mathrm y_{\iY+\bar m^{\mathrm y}}, \mathrm x^\mathrm y_{\iY+2\bar m^{\mathrm y}})$. 

\begin{table}[tb]
\centering
\begin{footnotesize}
\begin{tabular}{ll}
$x, y, z$ & spatial coordinate axes \\
$\R, \T$& reference, template image functions, $\R, \T: \rr^3 \to \rr$\\
$\omr$ & reference image domain, $\omr \subset \rr^3$ \\
$\trafo$& deformation function, $\trafo: \omr \to \rr^3$ \\
$m, \my$ & number of grid cells in the image/deformation grid, $m = \left(m_x,m_y,m_z\right)$\\
$\bar m, \barmy$ & total number of grid cells, $\bar m = m_x m_y m_z$\\
$h, \hy$ & grid spacings for each coordinate direction, $h = \left(h_x,h_y,h_z\right)$\\
$\bar h, \barhy$ & volume of a single grid cell, $\bar h = h_x h_y h_z$\\
$i$ & linear index with $i = \hat i + \hat j m_x + \hat k m_x m_y$ \\
$\iMinusX, \iPlusX$& indices of neighbors in $x$-direction, taking boundary conditions into account\\
$\iMinusY, \iPlusY$& indices of neighbors in $y$-direction, taking boundary conditions into account\\
$\iMinusZ, \iPlusZ$& indices of neighbors in $z$-direction, taking boundary conditions into account\\
$\vecx, \vecxy$ & vector of lexicographically ordered grid points, $\vecx = \left(\x_1,\ldots,\x_{3\bar m}\right) \in \rr^{3\bar m}$ \\
$\vecx_i, \vecxy_i$ & $i$-th grid point, $\vecx_i = \left(\x_i,\x_{i+\bar m},\x_{i+2\bar m}\right) \in \rr^3$ \\
$\vecy$ & deformation discretized on deformation grid, $\vecy = (\y_1,\ldots,\y_{3\barmy}) \in \rr^{3\barmy}$,
\\
&$\left(\y_i,\y_{i+\barmy},\y_{i+2\barmy}\right) = \trafo(\vecxy_i)$\\
$\vecu$ & displacement discretized on deformation grid, $\vecu = \vecy - \vecxy \in \rr^{3 \barmy}$  \\
$P$ & grid conversion operator from deformation grid to image grid,\\ & $P: \rr^{3\barmy} \to \rr^{3\bar m}$\\
$\hat \vecy$ & deformation discretized on image grid points, $\hat \vecy = P(\vecy) \in \rr^{3 \bar m}$ \\
$\hat \vecy_i$ & $i$-th deformation point, 
$\hat \vecy_i = \left( P(\vecy)_i, P(\vecy)_{i+\bar m}, P(\vecy)_{i+2\bar m}\right)$ \\
$R(\vecx)$ & reference image evaluated on image grid, $R:\rr^{3 \bar m} \to \rr^{\bar m}$ 
\\
$R_i$ &scalar value of reference image at $i$-th grid point, $R_i = \R(\vecx_i) \in \rr$ \\
$T(P(\vecy))$ & template evaluated on deformed image grid, $T:\rr^{3 \bar m} \to \rr^{\bar m}$
\\
$T_i$ &scalar value of reference image at deformed $i$-th grid point, $T_i = \T(\hat \vecy_i) \in \rr$\\
 \end{tabular}
 \end{footnotesize}
\caption{Summary of notation introduced in Section~\ref{sec:discretization} for discretization of the continuous registration problem. All notation regarding the deformation grid is identical to the corresponding notation for the image grid, except for an additional superscript ``$\y$'', e.g., $m$ denotes the number of grid cells in the image grid, and $m^y$ denotes the number of grid cells in the deformation grid.}
\label{table:notation}
\end{table}

We refer to~\cite{FAIR} for further discussion on different grids. In this work, the deformation grid size is always chosen with $m^{\mathrm y}_{\kY}-1 \leq m_k,\ k=x,y,z$, so that the image grid is as least as fine as the deformation grid.

The separate choice of grids allows to use a coarser grid for the deformation -- which directly affects the number of unknowns -- and a high-resolution grid for the input images. However, it adds an extra interpolation step, as in order to compute the distance measure, the deformed template image $\mathcal T(\varphi)$ needs to be evaluated on the same grid as the reference image $\mathcal R$.

We use a linear interpolation function  $P: \mathbb R^{3\bar m^{\mathrm y}} \to \mathbb R^{3\bar m}$, mapping between image grid and deformation grid, and define 
$T(P(\mathbf y)) := \left(\mathcal T(\mathbf{\hat y}_i)\right)_{i=1,\ldots,\bar m} \in \mathbb R^{\bar m}$ with $\mathbf{\hat y}_{i} := \left(P(\mathbf y)_i,P(\mathbf y)_{i+\bar m},P(\mathbf y)_{i+2\bar m}\right)$. 
For evaluating the template image $\mathcal T$, tri-linear interpolation with Dirichlet boundary conditions is used. This is a reasonable assumption for medical images, which often exhibit a black background.

Discretizing the integral in \eqref{eq:ngf} using the midpoint quadrature rule and denoting by $T_i, R_i$ the $i$-th component function, the compact expression
\begin{align}
 D(\mathbf{y}) = \bar h \sum_{i=1}^{\bar m} \left( 1 - \left( \frac{\frac12 \langle \tilde \nabla  T_i(P(\mathbf{y})),\tilde \nabla R_i(\mathbf{x})\rangle+\tau\varrho}{\Vert \tilde \nabla T_i(P(\mathbf{y})) \Vert_\tau \Vert \tilde \nabla R_i(\mathbf{x}) \Vert_\varrho}\right)^2 \right) \label{eq:NGF_discretized}
 \end{align}
 is obtained for the full discretized NGF term. The factor $1/2$ is caused by the ``square-then-average'' scheme for discretizing the gradient $\tilde \nabla$:
Given a discretized image $I \in \mathbb R^{\bar m}$, we define the \emph{backward difference operator} at the $i$-th grid point, $\tilde \nabla_- I_i: \mathbb R^{\bar m} \to \mathbb R^3$, as
\begin{align*}
\tilde \nabla_- I_i := \left(\frac{I_i - I_{\iMinusX}}{h_x},\frac{I_{i} - I_{\iMinusY}}{h_y},\frac{I_{i} - I_{\iMinusZ}}{h_z}\right),
\end{align*}
where the neighborhood is defined via
\begin{align}
 &\iMinusX := \max(\hat i - 1,1) + \hat j\, m_x + \hat k m_x m_y&& \iPlusX := \min(\hat i + 1,m_x) + \hat j\, m_x + \hat k m_x m_y \nonumber \\
 &\iMinusY := \hat i + \max(\hat j - 1,1) m_x + \hat k m_x m_y && \iPlusY := \hat i + \min(\hat j + 1,m_y)\, m_x + \hat k m_x m_y \label{eq:boundary_conditions}\\
 &\iMinusZ := \hat i + \hat j m_x + \max(\hat k - 1,1) m_x m_y && \iPlusZ := \hat i + \hat j m_x + \min(\hat k + 1,m_z)\, m_x m_y \nonumber 
\end{align}
and $i_0 := i$. This allows to implement Neumann boundary conditions without special treatment of the boundary cells. By $\tilde \nabla_+ I_i$ we denote the corresponding forward finite differences operator. We combine both operators into one, 
\begin{align}
\tilde \nabla I_i := \left( \tilde \nabla_- I_i , \tilde \nabla_+ I_i\right) \label{eq:gradient_discretized} 
\end{align}
with $\tilde \nabla I_i: \mathbb R^{\bar m} \to \mathbb R^6$,
and define $\Vert \tilde \nabla I_i \Vert_\varepsilon := \sqrt{\frac12 \langle \tilde \nabla I_i, \tilde \nabla I_i \rangle + \varepsilon^2}$. This ``square-then-average'' scheme for approximating the terms in \eqref{eq:ngf} allows to use ``short'' forward and backward differences, which better preserve high frequency gradients than ``long'' central differences~\cite{heldmann2006non}. The factor $1/2$ is required in order to faithfully discretize \eqref{eq:ngf}, and can be interpreted as an averaging of the squared norms of the forward and backward operators.

Similarly to the NGF, the curvature regularizer is discretized as
\begin{align}
 S(\mathbf{y}) = \bar h^\mathrm y \sum_{i=1}^{\bar m^\mathrm y} \sum_{d=0}^{2}\left( \tilde \Delta  \mathbf{u}_{i+d\bar m^\mathrm y}\right)^2, \label{eq:CURV_discretized}
\end{align}
with the discretized \emph{displacement} $\mathbf{u} := (\mathrm u_1, \ldots, \mathrm u_{3\bar m^\mathrm y} ) :=  \mathbf{y}-\mathbf x^\mathrm y$ and the discretized Laplace operator
\begin{align}
\tilde \Delta \mathbf{u}_{i+d\bar m^\mathrm y} := \sum_{k \in \{x,y,z\}} \frac{1}{\left(h^\mathrm y_k\right)^2}\left(
\mathrm{u}_{\iMinusK+d\bar m^\mathrm y} 
- \,2\, \mathrm{u}_{i+d\bar m^\mathrm y} \label{eq:laplace_discretized} 
+ \mathrm{u}_{\iPlusK+d\bar m^\mathrm y} \right)
\end{align}
with homogeneous Neumann boundary conditions.
Overall, we obtain the discretized version
\begin{align}
\min_{\mathbf{y} \in \mathbb R^{3 \bar m^\mathrm y}} J(\mathbf{y}), \quad J(\mathbf{y}) := D(\mathbf{y}) + \alpha S(\mathbf{y})\label{eq:discreteenergy}
\end{align}
of the minimization problem \eqref{eq:optimizationProblem}, which can then be solved using quasi-Newton methods, as summarized in the following section.

\subsection{Numerical optimization} \label{sec:numerical-optimization}

In order to find a minimizer of the discretized objective function \eqref{eq:discreteenergy}, we use iterative Newton-like optimization schemes. In each step of the iteration, an equation of the form
\begin{align}
 \hat \nabla^2 J(\mathbf{y}^k) \mathbf{s}^k = -\nabla J(\mathbf{y}^k) \label{eq:newton-equation}
\end{align}
is solved for a descent direction $\mathbf{s}^k$. Then $\mathbf{y}^k$ is updated via 
\begin{equation*}
\mathbf{y}^{k+1}  = \mathbf{y}^k + \eta \mathbf{s}^k,
\end{equation*} 
where the step length $\eta$ is determined by Armijo line search~\cite{nocedal1999numerical,FAIR}. The matrix $\hat \nabla^2 J$ should approximate the Hessian $\nabla^2 J(\mathbf{y}^k)$. Here we consider the \emph{Gauss-Newton} scheme and the \emph{L-BFGS} scheme, which have both been used in different image registration applications~\cite{vercauteren2009diffeomorphic, konig2015parallel, ruhaak2013highly}. 
The minimization is embedded in a coarse-to-fine multi-level scheme, where the problem is solved on consecutively finer deformation and image grids.

\subsubsection{Gauss-Newton} \label{sec:gauss-newton}
The Gauss-Newton scheme uses a quadratic approximation of the Hessian and is suitable for least-squares type objective functions of the form
\begin{align}
\hat J(\mathbf{y}) = \hat r (\mathbf{y})^\top \hat r (\mathbf{y})= \Vert \hat r (\mathbf{y}) \Vert_2^2,\label{eq:jyr}
\end{align}
where $\hat r(\mathbf{y})$ is a residual function, which can depend in a nonlinear way on the unknown~$\mathbf{y}$. 
The gradient in the Newton equation \eqref{eq:newton-equation} can be written as $\nabla \hat J = 2 \,\hat r^\top\, d \hat r$, where $d \hat r$ is the Jacobian of $\hat r$, and the Hessian is approximated by $H := \hat \nabla^2 \hat J = 2 d \hat r^\top d \hat r$~\cite{nocedal1999numerical}. 
This approximation discards second-order derivative parts of the Hessian and guarantees a symmetric positive semi-definite Hessian approximation $H$, so that $\mathbf{s}^k$ in~\eqref{eq:newton-equation} is always a descent direction. 

We apply this approximation only to the Hessian of the distance measure $D$,  $\nabla^2 D \approx H$. For the regularizer $S$, we use the exact Hessian, which is readily available, as $S$ is a quadratic function.
The (quasi\mbox{-})Newton-equation \eqref{eq:newton-equation} can be written as
\begin{align*}
(H + \alpha  \nabla^2 S) \mathbf{s}^k = -(\nabla D + \alpha \nabla S),
\end{align*}
which is then approximately solved in each step using a conjugate-gradient (CG) iterative solver~\cite{nocedal1999numerical}. 

\subsubsection{L-BFGS}
Instead of directly determining a Hessian approximation at each step, the L-BFGS scheme iteratively updates the approximation using previous and current gradient information.

L-BFGS requires an initial approximation $H_0$ of the Hessian, which is often chosen to be a multiple of the identity. However, in our case we can incorporate the known Hessian of the regularizer $S$ by choosing $H_0 = \nabla^2 S + \gamma I$, where $I$ is the identity matrix and $\gamma > 0$ is a parameter in order to make $H_0$ positive definite; further details can be found in~\cite{FAIR,nocedal1999numerical}. 

\section{Analytical derivatives}\label{sec:DeriComp}

As can be seen from Section~\ref{sec:numerical-optimization}, computing derivatives of distance measure and regularizer is a critically important task for the minimization of the discretized objective function. In this section, algebraic formulations of the required derivatives are presented, which allow for a closer analysis and construction of specialized computational schemes. 

\subsection{Curvature}

The discretized curvature regularizer \eqref{eq:CURV_discretized} is a quadratic function involving the (linear) Laplace operator $\tilde \Delta  \mathbf{u}_{i}$. Thus, using the chain rule
the $i$-th element of the gradient can explicitly be computed by
\begin{align}
 \left(\nabla S(\mathbf{y})\right)_{i+d\bar m^{\mathrm y}} = 2\, \bar h^\mathrm y \tilde \Delta \left(\tilde \Delta  \mathbf{u}\right)_{i + d\bar m^{\mathrm y}}, \label{eq:curvatureMF}
\end{align}
with $d \in \{0,1,2\}$ for the directional derivatives. The Hessian is constant and can be immediately seen from \eqref{eq:laplace_discretized}.

\subsection{NGF} \label{sec:NGF_derivative}

The inner computations of the NGF in \eqref{eq:NGF_discretized} can be rewritten more compactly by introducing a residual function $r: \mathbb R^{\bar m} \to \mathbb R^{\bar m}$ with components
\begin{align}
  r_i(T) :=  \frac{\frac12 \langle \tilde \nabla  T_i,\tilde \nabla R_i\rangle+\tau\varrho}{\Vert \tilde \nabla T_i \Vert_\tau \Vert \tilde \nabla R_i \Vert_\varrho}, \label{eq:residual}
\end{align}
so that the NGF term in \eqref{eq:NGF_discretized} becomes
\begin{align}
 D(\mathbf{y}) = \bar h \sum_{i=1}^{\bar m} \left( 1- r_i(T(P(\mathbf y)))^2\right) \label{eq:NGF_discretized_ri}.
\end{align}
Introducing a reduction function $\psi: \mathbb R^{\bar m} \to \mathbb R, \left(r_1,\ldots,r_{\bar m}\right)^\top \mapsto \bar h \sum_{i=1}^{\bar m} (1- r_i^2)$, it holds
\begin{align}
D(\mathbf y) = \psi(r(T(P(\mathbf y)))).
\end{align}
This composite function maps the deformation $\mathbf y \in \mathbb R^{3\bar m^{\mathrm y}}$ onto a scalar image similarity in four steps,
\begin{align}
 \mathbb R^{3\bar m^{\mathrm y}} \overset{P}{\rightarrow} \mathbb R^{3\bar m} \overset{T}{\rightarrow} \mathbb R^{\bar m} \overset{r}{\rightarrow} \mathbb R^{\bar m} \overset{\psi}{\rightarrow} \mathbb R. \label{eq:function_chain}
\end{align}
Using the chain rule, the gradient of $D$ can now be obtained as a product of (sparse) Jacobian matrices 
\begin{align}
 \nabla D = \left(\frac{\partial \psi}{\partial r} \frac{\partial r }{\partial T} \frac{\partial T }{\partial P} \frac{\partial P}{\partial \mathbf y}\right)^\top. \label{eq:NGF_discretized_gradient}
\end{align}
The Gauss-Newton approximation of the Hessian (Section~\ref{sec:gauss-newton}) becomes
\begin{align}
 H = 2\, \bar h\ \frac{\partial P}{\partial \mathbf y}^\top \frac{\partial T }{\partial P}^\top \frac{\partial r }{\partial T}^\top \frac{\partial r }{\partial T} \frac{\partial T }{\partial P} \frac{\partial P}{\partial \mathbf y}. \label{eq:NGF_discretized_hessian}
\end{align}
Note that, in contrast to the classical Gauss-Newton method \eqref{eq:jyr}, the residual in \eqref{eq:NGF_discretized_ri} has a different sign. Therefore, in order to compute a descent direction in~\eqref{eq:newton-equation}, the sign of the Hessian approximation has been inverted (see also \cite{FAIR}).

Using these formulations, the evaluation of the gradient and Hessian can be implemented based on sparse matrix representations~\cite{FAIR}. However, while insightful for educational and analytic purposes, these schemes have important shortcomings: storing the matrix elements requires large amounts of memory, while assembling the sparse matrices and sparse matrix multiplications  impede efficient parallelization. In the following chapters, we will therefore focus on strategies for direct evaluation that do not require intermediate storage and are highly parallel. 

\section{Derivative analysis and matrix-free computation schemes}\label{sec:MatrixFree}
From a computational perspective, the NGF \eqref{eq:NGF_discretized_ri} and curvature formulations \eqref{eq:CURV_discretized} are very amenable to parallelization, as all summands are independent of each other, and there are no inherent intermediate matrix structures required. However, this advantage is lost when the evaluation of the derivatives is implemented using individual matrices for the factors in \eqref{eq:NGF_discretized_gradient} and \eqref{eq:NGF_discretized_hessian}.

In order to exploit the structure of the partial derivatives, in the following sections we will analyze the matrix structure of all derivatives and derive equivalent closed-form expressions that do not rely on intermediate storage.

We will make substantial use of the fact that the matrix structure of the partial derivatives is independent of the data. Therefore, the locations of non-zero elements are known a priori, which allows to derive efficient sparse schemes.

\subsection{Derivative computations for NGF} \label{sec:mf_methods_ngf}

\subsubsection{Gradient} \label{sec:mf_methods_ngf_gradient}

The NGF gradient computation in \eqref{eq:NGF_discretized_gradient} involves four different Jacobian matrices. The derivative of the vector reduction $\psi(r)$ is straightforward:
\begin{align}
 \frac{\partial \psi}{\partial r} = - 2 \bar h r  = -2 \bar h \left(r_1,\ldots,r_{\bar m}\right) \in \mathbb R^{1 \times \bar m}. \label{eq:dPsi}
\end{align}
In contrast, the computation of $\frac{\partial r}{\partial T}  \in \mathbb R^{\bar m \times \bar m}$ is considerably more involved. In order to calculate a single row of this Jacobian $\frac{\partial r_i}{\partial T}$, the definition \eqref{eq:NGF_discretized_ri} can be used and interpreted as a quotient 
\begin{align}
 r_i = \frac{r_{1,i}}{r_{2,i}} =  \frac{\frac12 \langle \tilde \nabla  T_i,\tilde \nabla R_i\rangle+\tau\varrho}{\Vert \tilde \nabla T_i \Vert_\tau \Vert \tilde \nabla R_i \Vert_\varrho}. \label{eq:ri_quotient}
\end{align}
Separately differentiating numerator and denominator, this yields 
\begin{align*}
&\frac{\partial r_{1,i}}{\partial T} = \frac12 (\tilde \nabla R_i)^\top \frac{\partial \tilde \nabla_i }{\partial T} \in \mathbb R^{1 \times \bar m} &&\text{and} &\frac{\partial r_{2,i}}{\partial T} = \Vert \tilde \nabla R_i \Vert_\varrho \frac{(\tilde \nabla T_i)^\top}{ 2 \Vert \tilde \nabla T_i \Vert_\tau} \frac{\partial \tilde \nabla_i }{\partial T} \in \mathbb R^{1 \times \bar m}.
\end{align*}

Included in both of these terms is the derivative of the finite difference gradient computation $\tilde \nabla T_i: \mathbb R^{\bar m} \to \mathbb R^6$ as defined in \eqref{eq:gradient_discretized}, i.e., mapping an image to the forward- and backward difference gradients.  As the gradient at a single point only depends on the values at the point itself and at neighboring points, the Jacobian $\frac{\partial \tilde \nabla_i }{\partial T} \in \mathbb R^{6 \times \bar m}$ is of the form

\begin{align*}
\frac{\partial \tilde \nabla_i }{\partial T} = 
\bordermatrix{
   &\text{\scriptsize{$\iMinusZ$}}&&\text{\scriptsize{$\iMinusY$}}&&\text{\scriptsize{$\iMinusX$}}&&\text{\scriptsize{$i$}}&&\text{\scriptsize{$\iPlusX$}}&&\text{\scriptsize{$\iPlusY$}}&&\text{\scriptsize{$\iPlusZ$}}  \cr
                  &              &\quad &              &\quad& \frac{-1}{h_x} &\quad&\frac{1}{h_x} &\quad &              &\quad&             &\quad&\cr
                  &              &      &\frac{-1}{h_y}&     &                &     &\frac{1}{h_y} &      &              &     &             &     &\cr
                  &\frac{-1}{h_z}&      &              &     &                &     &\frac{1}{h_z} &      &              &     &             &     &\cr
                  &              &      &              &     &                &     &\frac{-1}{h_x}&      &\frac{1}{h_x} &     &             &     &\cr
                  &              &      &              &     &                &     &\frac{-1}{h_y}&      &              &     &\frac{1}{h_y}&     &\cr
                  &              &      &              &     &                &     &\frac{-1}{h_z}&      &              &     &             &     &\frac{1}{h_z}\cr
  },
\end{align*}  
where only non-zero elements are shown and the non-zero column indices are displayed above the matrix. 
The quotient rule, applied to \eqref{eq:ri_quotient}, 
leads to
\begin{align}
 \frac{\partial r_i}{\partial T} &= \frac{1}{r_{2,i}^2}\left( \frac{\partial r_{1,i}}{\partial T}  r_{2,i} - r_{1,i} \frac{\partial r_{2,i}}{\partial T}\right) =  \frac{\partial r_{1,i}}{\partial T}  \frac{1}{r_{2,i}} - \frac{r_{1,i}}{r_{2,i}^2} \frac{\partial r_{2,i}}{\partial T}\\
 &= \frac{(\tilde \nabla R_i)^\top}{2\Vert \tilde \nabla T_i \Vert_\tau \Vert \tilde \nabla R_i \Vert_\varrho}  \frac{\partial \tilde \nabla_i }{\partial T}  - \frac{\frac12 \langle \tilde \nabla  T_i,\tilde \nabla R_i\rangle+\tau\varrho}{\Vert \tilde \nabla T_i \Vert_\tau^2 \Vert \tilde \nabla R_i \Vert_\varrho^2} \Vert \tilde \nabla R_i \Vert_\varrho \frac{(\tilde \nabla T_i)^\top}{ 2 \Vert \tilde \nabla T_i \Vert_\tau} \frac{\partial \tilde \nabla_i }{\partial T} \\
 &= \frac12 \left(  \frac{(\tilde \nabla R_i)^\top}{\Vert \tilde \nabla T_i \Vert_\tau \Vert \tilde \nabla R_i \Vert_\varrho} \frac{\partial \tilde \nabla_i }{\partial T}  - \frac{\frac12 \langle \tilde \nabla  T_i,\tilde \nabla R_i\rangle+\tau\varrho}{\Vert \tilde \nabla T_i \Vert_\tau^3 \Vert \tilde \nabla R_i \Vert_\varrho} (\tilde \nabla T_i)^\top \frac{\partial \tilde \nabla_i }{\partial T}\right).\label{eq:ritcomp}
\end{align}
We introduce the abbreviation 
\begin{align}
 \rho_i(k) := \frac{-R_{i} + R_{\iPlusK}}{\Vert \tilde \nabla T_i \Vert_\tau \Vert \tilde \nabla R_i \Vert_\varrho} - \frac{\left(\frac12 \langle \tilde \nabla  T_i,\tilde \nabla R_i\rangle+\tau\varrho\right)\left(-T_{i} + T_{\iPlusK}\right)}{\Vert \tilde \nabla T_i \Vert_\tau^3 \Vert \tilde \nabla R_i \Vert_\varrho}, \label{eq:rho}
\end{align}
and the set of indices of non-zero elements
\begin{align}
\mathcal K := \{ -z, -y, -x,\ 0,x,y,z \}. \label{eq:k}
\end{align} 
Furthermore, define $\hat h_k := \frac{1}{2\left( h_{\vert k \vert}\right)^2}$ and
\begin{align}
 \hat \rho_i(k) := 
\begin{cases}
\sum_{j \in \mathcal K \setminus \{0\}} -\hat h_{j} \rho_i(j)&\text{if}\ k=0,  \\ 
 \hat h_k \rho_i(k) &\text{otherwise.}
\end{cases}\label{eq:hatRho}
\end{align}
Then, \eqref{eq:ritcomp} can be compactly written as
\begin{align}
 \frac{\partial r_i}{\partial T} &= 
  \bordermatrix{
           & \text{\scriptsize{$\iMinusZ$}}&\text{\scriptsize{$\iMinusY$}}&\text{\scriptsize{$\iMinusX$}}&\text{\scriptsize{$i$}}&\text{\scriptsize{$\iPlusX$}}&\text{\scriptsize{$\iPlusY$}}&\text{\scriptsize{$\iPlusZ$}} \cr
                                    &\hat \rho_i(-z)&
                                     \hat \rho_i(-y) &
                                     \hat \rho_i(-x) &
                                     \hat \rho_i(0) &
                                     \hat \rho_i(x) &
                                     \hat \rho_i(y) &
                                     \hat \rho_i(z) 
                                   } \in \mathbb R^{1 \times \bar m},   \label{eq:dr} 
\end{align}
where only non-zero elements are shown and single element locations are denoted above the vector. It can be seen that \eqref{eq:dr} exhibits a very sparse pattern with only seven non-zero elements.

For a full gradient computation as in \eqref{eq:NGF_discretized_gradient}, it remains to consider the terms $\frac{\partial T}{\partial P}$ and~$\frac{\partial P}{\partial \mathbf y}$. The first term represents the derivative of the template image interpolation function with respect to the image grid coordinates. Given the lexicographical ordering of the grid points, the image derivative matrix is composed of three diagonal matrices,
\begin{align}
               \frac{\partial T}{\partial P} = \begin{pmatrix}
                  \mathrm{diag}\left(\frac{\partial T_1}{\partial P_1} \ldots \frac{\partial T_{\bar m}}{\partial P_{\bar m}}\right),  
                  \mathrm{diag}\left(\frac{\partial T_1}{\partial P_{\bar m+1}} \ldots \frac{\partial T_{\bar m}}{\partial P_{2 \bar m}}\right),
                  \mathrm{diag}\left(\frac{\partial T_1}{\partial P_{2 \bar m+1}} \ldots \frac{\partial T_{\bar m}}{\partial P_{3\bar m}}\right)
                \end{pmatrix},  \label{eq:dT}
\end{align}
where each non-zero element represents the derivative at a single point with respect to a single coordinate.

The last remaining term $\frac{\partial P}{\partial \mathbf y}$ is the Jacobian of the grid conversion function. This is a linear operator, which will be analyzed in detail in Section~\ref{sec:grid-conversion}.

From the derivations in \eqref{eq:dPsi}, \eqref{eq:dr} and \eqref{eq:dT}, it can be seen that the main components of the NGF gradient computation exhibit a sparse structure with fixed patterns.
The main effort comes from computing the derivative of~$r$, which has a seven-banded diagonal structure. 

Exploiting this pattern, a single element of the partial gradient $\frac{\partial D}{\partial P} := \frac{\partial \psi}{\partial r} \frac{\partial r }{\partial T} \frac{\partial T }{\partial P} \in \mathbb R^{1 \times 3 \bar m}$
can be explicitly computed as 
\begin{align}
 \left(\frac{\partial D}{\partial P}\right)_{i+d\bar m} 
 &= -2\, \bar h \left( 
 \sum_{k \in \mathcal K} r_{\iPlusK}  \hat \rho_{\iPlusK} (-k)\right) \frac{\partial T_i}{\partial P_{i+d \bar m} },
 \label{eq:gradNGF}
\end{align}
where $d \in \{0,1,2\}$ and $\mathcal K = \{ -z, -y, -x,\ 0,x,y,z \}$ as in \eqref{eq:k}. Algorithm~\ref{alg:grad_NGF} shows a pseudocode implementation for computing the individual elements of the gradient as in \eqref{eq:gradNGF}.

This formulation has multiple benefits:
Any element of the gradient can be computed directly from the input data, without the need for (sparse) matrices and without having to store intermediate results. Furthermore, gradient elements can be computed independently, which allows for a fully parallel implementation. As will be discussed in Section~\ref{sec:algorithm_analysis}, both of these properties substantially decrease memory usage and  computation time.

\begin{algorithm}[t]
\renewcommand{\algorithmicensure}{\textbf{Output:}}
\begin{algorithmic}[1]
 \For{$\hat k \textup{ in } [0, m_z-1] $}
   \For{$\hat j \textup{ in } [0, m_y-1] $}
     \For{$\hat i \textup{ in } [0, m_x-1] $}
       \State{$i \gets \hat i + m_x \hat j + m_x m_y \hat k$}  \Comment{Compute linear index}
       \State{$i \pm x, i \pm y, i\pm z \gets \text{as in \eqref{eq:boundary_conditions}}$} \Comment{Compute neighbor indices}
       \State{$[\texttt{dTx},\texttt{dTy},\texttt{dTz}] \gets \texttt{imageDerivative}(T_i)$}  \Comment{Compute image derivative}
       \State{}
       \State{$\texttt{r} \gets [r_{\iMinusZ},r_{\iMinusY},r_{\iMinusX},r_{i},r_{\iPlusX},r_{\iPlusY},r_{\iPlusZ}]$} \Comment{$r_i$ as defined in \eqref{eq:ri_quotient}}
       \State{$\texttt{dr} \gets [\hat\rho_{\iMinusZ}(z),\hat\rho_{\iMinusY}(y),\hat\rho_{\iMinusX}(x),\hat\rho_{i}(0),\hat\rho_{\iPlusX}(-x),\hat\rho_{\iPlusY}(-y),\hat\rho_{\iPlusZ}(-z)]$} 
       \State{} \Comment{$\hat\rho_i$ as defined in \eqref{eq:hatRho}}
       
       \State{$\texttt{drSum} \gets -2\bar h\,(\texttt{r[0]dr[0]}+\texttt{r[1]dr[1]}+\texttt{r[2]dr[2]}+\texttt{r[3]dr[3]}$} 
       \State{\hspace*{2.0cm}$+\ \texttt{r[4]dr[4]}+\texttt{r[5]dr[5]}+\texttt{r[6]dr[6]})$}
       \State{}\Comment{Compute sum of \eqref{eq:gradNGF}}      
       \State{$\texttt{grad[$i$\hphantom{${}+ 2 \bar m$}]} \gets \texttt{drSum} \cdot \texttt{dTx}$}
       \State{$\texttt{grad[$i+ \hphantom{2} \bar m $]} \gets \texttt{drSum} \cdot \texttt{dTy}$}
       \State{$\texttt{grad[$i+ 2 \bar m$]} \gets \texttt{drSum} \cdot \texttt{dTz}$}
       
     \EndFor
   \EndFor
 \EndFor
  \Ensure {$\texttt{grad[$i+ d \bar m$]} = \left(\frac{\partial D}{\partial P}\right)_{i+d\bar m}$}
\end{algorithmic}
\caption{Pseudocode for matrix-free computation of the elements of the NGF gradient as in \eqref{eq:gradNGF}. The algorithm can be fully parallelized over all loop iterations.}\label{alg:grad_NGF}
\end{algorithm}

\subsubsection{Hessian-vector multiplication} \label{sec:hessian-vector_multiplication}

As noted in Section~\ref{sec:gauss-newton}, when using the Gauss-Newton scheme for optimization, linear systems involving the quadratic approximation $H \in \mathbb R^{3 \bar m^{\mathrm y} \times 3 \bar m^{\mathrm y}}$ of the Hessian \eqref{eq:NGF_discretized_hessian} need to be solved in each iteration. Although $H$ is sparse, the memory requirements for storing the final as well as intermediate matrices can still be considerable.
Therefore, in the following we will consider an efficient scheme for evaluating the matrix-vector product $H \mathbf p = \mathbf q$ with $\mathbf{p}, \mathbf{q} \in \mathbb R^{3 \bar m^{\mathrm y}}$, which is the foundation for using iterative methods, such as conjugate gradients, in order to solve the (Gauss-)Newton equation \eqref{eq:newton-equation}.

Figure~\ref{fig:hess_mult_schema} shows a schematic of the computations involved. Again, we consider the Jacobian $\frac{\partial P}{\partial \mathbf y}$ and its transpose in \eqref{eq:NGF_discretized_hessian} as separate grid conversion steps, which will be discussed in detail in Section~\ref{sec:grid-conversion}. The main computations consist of computing the matrix product
\begin{align}
 \hat H := \frac{\partial T}{\partial P}^\top \frac{\partial r}{\partial T}^\top \frac{\partial r}{\partial T} \frac{\partial T}{\partial P} \in \mathbb R^{3 \bar m \times 3 \bar m}, \label{eq:HNGF}
\end{align}
with the components $\frac{\partial T}{\partial P} \in \mathbb R^{ \bar m \times 3 \bar m}$, $\frac{\partial r}{\partial T} \in \mathbb R^{\bar m \times \bar m}$, which is equivalent to computing the approximate Hessian $H$ in \eqref{eq:NGF_discretized_hessian} with the exception of the grid conversion steps.

We first analyze the matrix-vector product $\hat H \mathbf{\hat p} = \mathbf{\hat q}$ with $\mathbf{\hat p}, \mathbf{\hat q} \in \mathbb R^{3 \bar m}$. Abbreviating $dr := \frac{\partial r}{\partial T}$, the main challenge is efficiently computing the matrix product $dr^\top dr$.

Using the definition and notation of a single \emph{row} of $dr$ given in \eqref{eq:dr}, a single \emph{column} of the matrix can be written as
\begin{align}
 \frac{\partial r}{\partial T_i} &= 
  \bordermatrix{
           & \text{\scriptsize{$\iMinusZ$}}&\text{\scriptsize{$\iMinusY$}}&\text{\scriptsize{$\iMinusX$}}&\text{\scriptsize{$i$}}&\text{\scriptsize{$\iPlusX$}}&\text{\scriptsize{$\iPlusY$}}&\text{\scriptsize{$\iPlusZ$}} \cr
                                    &\hat \rho_{\iMinusZ}(z)&
                                     \hat \rho_{\iMinusY}(y) &
                                     \hat \rho_{\iMinusX}(x) &
                                     \hat \rho_i(0) &
                                     \hat \rho_{\iPlusX}(-x) &
                                     \hat \rho_{\iPlusY}(-y) &
                                     \hat \rho_{\iPlusZ}(-z) 
                                   }\mathclose{\vphantom{\Big)}}^\top,\label{eq:drT}   
\end{align}
again only showing non-zero elements at the indices denoted above the vector. As in \eqref{eq:dr}, a single column contains only seven non-zero elements.

Abbreviating $dr_i := \frac{\partial r}{\partial T_i} \in \mathbb R^{\bar m \times 1}$, each element of the matrix  $dr^\top dr$ is a scalar product of columns $\langle dr_i, dr_j \rangle$ for $i, j = 1,\ldots,\bar m$. However, due to the sparsity of the columns of $dr$ in \eqref{eq:drT}, there are only few non-zero scalar products. As a basic example,
consider the diagonal elements of $dr^\top dr$. In column $dr_i$, non-zero elements are located at $\left(dr_i\right)_{\iPlusK}, k\in \mathcal K$, with $\mathcal K$ as in \eqref{eq:k}.
This yields the scalar products $\langle dr_i, dr_i\rangle =\sum_{k \in \mathcal K} (dr_i)_{\iPlusK}^2 = \sum_{k \in \mathcal K} \hat \rho_{\iPlusK}(-k)^2$, a sum of seven terms.

More generally, for arbitrary $i,j$, define  $\kappa := j-i$, so that $\langle dr_i, dr_j \rangle = \langle dr_i, dr_{i+\kappa} \rangle$.
In order to characterize the elements where $\langle dr_i, dr_{i+\kappa} \rangle$ can be non-zero, observe that the non-zero elements of $dr_i$ are located at indices $i + \mathcal{M}$,
\begin{align}
 \mathcal M := \{ -m_x m_y, -m_x, -1,\ 0,1,m_x,m_x m_y \}.
\end{align}
Consequently, $\langle dr_i, dr_{i+\kappa} \rangle$ can only be non-zero if $(i+\mathcal M) \cap (i+\kappa+\mathcal{M})\neq\emptyset$, i.e., if the scalar product involves at least two non-zero elements. Equivalently, $\kappa \in \mathcal{N} := \mathcal{M} - \mathcal{M}$, therefore the number of non-zero inner products is bounded from above by the number of elements $|\mathcal{N}|$ in $\mathcal{N}$. Naturally, $|\mathcal{N}| \leq |\mathcal{M}|\cdot |\mathcal{M}|=49$, however it turns out that this bound can be substantially lowered.

In order to do so, define the mapping
  \begin{align*}
 N(i,j) := j-i, &&
 N: \mathcal M \times \mathcal M \to \mathbb Z ,
\end{align*}
so that 
\begin{align*}
\mathcal N = N(\mathcal M, \mathcal M) = \Big\{ \kappa \in \mathbb Z  \ \Big\vert\ \kappa = N(i,j),\ i,j \in \mathcal M\Big\}.
\end{align*}
Inspecting all possible combinations for $i$ and $j$, it turns out that $|\mathcal{N}| = 25$ (Table~\ref{tab:hessian_indices}),
which implies that each column of $dr^\top dr$ has at most 25 non-zero elements.

\begin{figure}[tb]
\centering
\begin{tikzpicture}[scale=1]

\node at (-2.8,1.8) [below] {$\frac{\partial P}{\partial \mathbf y}^\top\, \cdot$};

\node at (-1.5,1.5) [scale=0.3]{\setlength{\arraycolsep}{4pt} $\begin{pmatrix}
  \bullet\\
  &\bullet\\
  &&\bullet\\
  &&&\bullet\\
  &&&&\bullet\\
  &&&&&\bullet\\
  &&&&&&\bullet\\
  &&&&&&&\bullet\\
  &&&&&&&&\bullet\\
  &&&&&&&&&\bullet\\
    \bullet\\
  &\bullet\\
  &&\bullet\\
  &&&\bullet\\
  &&&&\bullet\\
  &&&&&\bullet\\
  &&&&&&\bullet\\
  &&&&&&&\bullet\\
  &&&&&&&&\bullet\\
  &&&&&&&&&\bullet\\
    \bullet\\
  &\bullet\\
  &&\bullet\\
  &&&\bullet\\
  &&&&\bullet\\
  &&&&&\bullet\\
  &&&&&&\bullet\\
  &&&&&&&\bullet\\
  &&&&&&&&\bullet\\
  &&&&&&&&&\bullet\\
 \end{pmatrix}$};

 \node at (-1.4,-0.5) [below] {$\frac{\partial T}{\partial P}^\top$};

\node at (0,1.5) [scale=0.3]{\setlength{\arraycolsep}{4pt} $\begin{pmatrix}
  \bullet&{\bullet}&&{\bullet}&&&{\bullet}\\
  {\bullet}&\bullet&{\bullet}&&{\bullet}&&&{\bullet} \\
  & {\bullet}&\bullet&{\bullet}&&{\bullet}&&&{\bullet} \\
  {\bullet}&& {\bullet}&\bullet&{\bullet}&&{\bullet}&&&{\bullet} \\
  &{\bullet}&& {\bullet}&\bullet&{\bullet}&&{\bullet}& \\
  &&{\bullet}&&{\bullet}&\bullet&{\bullet}&&{\bullet}& \\
  {\bullet}&&&{\bullet}&&{\bullet}&\bullet&{\bullet}&&{\bullet} \\
  &{\bullet}&&&{\bullet}&&{\bullet}&\bullet&{\bullet}& \\
  &&{\bullet}&&&{\bullet}&&{\bullet}&\bullet&{\bullet} \\
  &&&{\bullet}&&&{\bullet}&&{\bullet}&\bullet \\
 \end{pmatrix}$};

 \node at (0,0.9) [below] {$\frac{\partial r}{\partial T}^\top$};

 \node at (1.5,1.5) [scale=0.3]{\setlength{\arraycolsep}{4pt} $\begin{pmatrix}
  \bullet&{\bullet}&&{\bullet}&&&{\bullet}\\
  {\bullet}&\bullet&{\bullet}&&{\bullet}&&&{\bullet} \\
  & {\bullet}&\bullet&{\bullet}&&{\bullet}&&&{\bullet} \\
  {\bullet}&& {\bullet}&\bullet&{\bullet}&&{\bullet}&&&{\bullet} \\
  &{\bullet}&& {\bullet}&\bullet&{\bullet}&&{\bullet}& \\
  &&{\bullet}&&{\bullet}&\bullet&{\bullet}&&{\bullet}& \\
  {\bullet}&&&{\bullet}&&{\bullet}&\bullet&{\bullet}&&{\bullet} \\
  &{\bullet}&&&{\bullet}&&{\bullet}&\bullet&{\bullet}& \\
  &&{\bullet}&&&{\bullet}&&{\bullet}&\bullet&{\bullet} \\
  &&&{\bullet}&&&{\bullet}&&{\bullet}&\bullet \\
 \end{pmatrix}$};

 \node at (1.5,0.8) [below] {$\frac{\partial r}{\partial T}$};

 \node at (4.4,1.5) [scale=0.3]{\setlength{\arraycolsep}{4pt} $\begin{pmatrix}
  \bullet&&&&&&&&&&\bullet&&&&&&&&&&\bullet\\
  &\bullet&&&&&&&&&&\bullet&&&&&&&&&&\bullet\\
  &&\bullet&&&&&&&&&&\bullet&&&&&&&&&&\bullet\\
  &&&\bullet&&&&&&&&&&\bullet&&&&&&&&&&\bullet\\
  &&&&\bullet&&&&&&&&&&\bullet&&&&&&&&&&\bullet\\
  &&&&&\bullet&&&&&&&&&&\bullet&&&&&&&&&&\bullet\\
  &&&&&&\bullet&&&&&&&&&&\bullet&&&&&&&&&&\bullet\\
  &&&&&&&\bullet&&&&&&&&&&\bullet&&&&&&&&&&\bullet\\
  &&&&&&&&\bullet&&&&&&&&&&\bullet&&&&&&&&&&\bullet\\
  &&&&&&&&&\bullet&&&&&&&&&&\bullet&&&&&&&&&&\bullet\\
 \end{pmatrix}$};

 \node at (4.4,0.8) [below] {$\frac{\partial T}{\partial P}$};

 \node at (7,1.8) [below] {$\cdot\,  \frac{\partial P}{\partial \mathbf y}\, \cdot$};

  \node at (7.65,1.5) [scale=0.3]{\setlength{\arraycolsep}{4pt} $\begin{pmatrix}
 &\bullet&\\
 &\bullet\\
 &\bullet\\
 &  \bullet\\
 &  \bullet\\
 &  \bullet\\
 &  \bullet\\
 &  \bullet\\
 &  \bullet\\
 &  \bullet\\ 
  &\bullet&\\
 &\bullet\\
 &\bullet\\
 &  \bullet\\
 &  \bullet\\
 &  \bullet\\
 &  \bullet\\
 &  \bullet\\
 &  \bullet\\
 &  \bullet\\
  &\bullet&\\
 &\bullet\\
 &\bullet\\
 &  \bullet\\
 &  \bullet\\
 &  \bullet\\
 &  \bullet\\
 &  \bullet\\
 &  \bullet\\
 &  \bullet\\
 \end{pmatrix}$};

 \node at (7.65,-0.5) [below] {$\mathbf p$};
 
 \node at (8.15,1.65) [below] {$=$};
 
   \node at (8.6,1.5) [scale=0.3]{\setlength{\arraycolsep}{4pt} $\begin{pmatrix}
 &\bullet&\\
 &\bullet\\
 &\bullet\\
 &  \bullet\\
 &  \bullet\\
 &  \bullet\\
 &  \bullet\\
 &  \bullet\\
 &  \bullet\\
 &  \bullet\\ 
  &\bullet&\\
 &\bullet\\
 &\bullet\\
 &  \bullet\\
 &  \bullet\\
 &  \bullet\\
 &  \bullet\\
 &  \bullet\\
 &  \bullet\\
 &  \bullet\\
  &\bullet&\\
 &\bullet\\
 &\bullet\\
 &  \bullet\\
 &  \bullet\\
 &  \bullet\\
 &  \bullet\\
 &  \bullet\\
 &  \bullet\\
 &  \bullet\\
 \end{pmatrix}$};

 \node at (8.6,-0.5) [below] {$\mathbf q$};

\end{tikzpicture}
\caption{Schematic view of Hessian-vector multiplication $H \mathbf p = \mathbf q$. The computation involves highly sparse matrices with a fixed pattern of non-zeros. The main computational effort results from the matrix multiplication $\frac{\partial r}{\partial T}^\top \frac{\partial r}{\partial T}$, where $\frac{\partial r}{\partial T}$ has seven non-zero diagonals.}\label{fig:hess_mult_schema}
\end{figure}

For a given offset $\kappa = j - i$, the pre-image $N^{-1}(\kappa)$ indicates the indices of non-zero elements contributing to the inner product $\langle dr_i, dr_{j} \rangle$.
As shown in the rightmost column in Table~\ref{tab:hessian_indices}, for the main diagonal of $dr^\top dr$, seven elements are involved in the scalar product, while for all other elements only either one or two products need to be computed. Using $\kappa = j-i$ and $(\hat i, \hat j) \in  N^{-1}(\kappa)$, it holds 
\begin{align}
 \left(dr^\top dr\right)_{i,j} &= \begin{cases}
                                  \displaystyle \sum_{(\hat i, \hat j) \in N^{-1}(j-i)}\left(dr_i\right)_{i+{\hat i}} \left(dr_{j}\right)_{j+{\hat j}}, &\text{if}\ j-i \in \mathcal N, \\
                                  0, &\text{otherwise}, \nonumber
                                 \end{cases}  
                                 \\
                               & = \begin{cases}
                                  \displaystyle \sum_{(\hat i, \hat j) \in N^{-1}(j-i)}  \hat \rho_{i+{\hat i}}(-\hat i)\ \hat \rho_{j+{\hat j}}(-\hat j), &\text{if}\ j-i \in \mathcal N, \\
                                  0, &\text{otherwise}, \label{eq:drTdr}
                                 \end{cases}
\end{align}
where, in a slight abuse of notation, $\hat \rho(-m_1 m_2)$ should be interpreted as $\hat \rho(-z)$, etc. 

With this and the explicit formulation of $\hat \rho_i(k)$ in \eqref{eq:rho} and \eqref{eq:hatRho}, the computational cost of calculating $dr^\top dr$ can be reduced substantially: multiplications involving zero elements are no longer considered, 
administrative costs for locating and handling sparse matrix elements have been eliminated, and only essential calculations remain.

Substituting \eqref{eq:drTdr} into the expression \eqref{eq:HNGF} for evaluating the matrix-vector product $\mathbf{\hat q}= \hat H \mathbf{\hat p}$, we obtain
\begin{align}
 \hat{\mathrm q}_{d\bar m+i} =  \sum_{\kappa \in \mathcal N} \sum_{l \in \{0,1,2\}} \frac{\partial T_i}{\partial P_{d\bar m+i}} \left( dr^\top dr \right)_{i,i+\kappa} \frac{\partial T_\kappa}{\partial P_{l \bar m+i+\kappa}}\ \hat{\mathrm p}_{l \bar m+i+\kappa}, \label{eq:hessMult}
\end{align}
for $d \in \{0,1,2\}$, using the definition of $\left( dr^\top dr\right)_{i,i+\kappa}$ from \eqref{eq:drTdr}.

Similar to the matrix-free gradient formulation in \eqref{eq:gradNGF}, this allows a direct computation of each element of the result vector $\mathbf{\hat q}= \hat H \mathbf{\hat p}$ fully in parallel and directly from the input data. An implementation in pseudocode is shown in Algorithm~\ref{alg:hess_NGF}.

\begin{table}[t]
\centering
\begin{small}
\begin{tabular}{llll}
  $ \kappa \in \mathcal N $  &  $ N^{-1}(\kappa)$ & & $|N^{-1}(\kappa)|$\\\hline
  $ -2m_1 m_2 $  & $\{(m_1 m_2,-m_1 m_2)\}$ & & 1  \\
  $ -m_1 m_2-m_1$  & $\{(m_1,-m_1 m_2)$,& $(m_1 m_2,-m_1)\}$ & 2\\
  $ -m_1 m_2-1$  & $\{(1,-m_1 m_2)$,& $(m_1 m_2,-1)\}$ & 2\\
  $ -m_1 m_2  $  & $\{(0,-m_1 m_2)$,& $(m_1 m_2,0)\}$ & 2\\
  $ -m_1 m_2+1$  & $\{(-1,-m_1 m_2)$,&$(m_1 m_2,1)\}$ & 2\\
  $ -m_1 m_2+m_1$  & $\{(-m_1,-m_1 m_2)$,&$(m_1 m_2,m_1)\}$ & 2\\
  $ -2m_1 $   & $\{(m_1,-m_1)$ & & 1\\
  $ -m_1-1$   & $\{(1,-m_1)$,&$(m_1,-1)\}$ & 2\\
  $ -m_1  $   & $\{(0,-m_1)$,&$(m_1,0)\}$ & 2\\
  $ -m_1+1$   & $\{(-1,-m_1)$,&$(m_1,1)\}$ & 2\\
  $ -2 $    & $\{(1,-1)\}$ & & 1\\
  $ -1 $    & $\{(0,-1)$,&$(1,0)\}$ & 2\\
  $  0 $    & $\{(\hat i,\hat i)\,|\,\hat i\in \mathcal{M}\}$ & & 7  \\
  $ 1 $    & $\{(0,1)$,&$(-1,0)\}$ & 2 \\
  $\vdots$ & $\vdots$ & $\vdots$ & \vdots
\end{tabular}
\end{small}
\caption{
Offsets $\kappa:=j-i$, for which $\langle dr_i, dr_j \rangle \neq 0$, corresponding to non-zero elements  $(dr^\top dr)_{i,j}$ in the matrix-product $dr^\top dr$. The pre-image sets $N^{-1}(\kappa)$ characterize the locations of the non-zero element in $dr_i$ and $dr_j$. The $11$ omitted cases are identical to the ones shown except for opposite sign. The rightmost column shows the number of products involving non-zero coefficients in the evaluation of  $\langle dr_i, dr_j \rangle$ and sums to $|\mathcal{M}|\cdot |\mathcal{M}| = 49$. 
} \label{tab:hessian_indices}
\end{table} 

\begin{algorithm}[tbh]
\renewcommand{\algorithmicensure}{\textbf{Output:}}
\begin{algorithmic}[1]

 \State{$\texttt{N} \gets\ $\scriptsize{$[-2 m_1 m_2, -m_1 m_2 - m_1, -m_1 m_2-1, -m_1 m_2, -m_1 m_2+1, -m_1 m_2+m_1, -2m_1, -m_1-1, -m_1,$}}
 \State{$\hphantom{\texttt{N} \gets\ \  } $\scriptsize{$-m_1+1, -2, -1, 0, 1, 2, m_1-1, m_1, m_1+1, 2m_1, m_1 m_2 - m_1, m_1 m_2 - 1, m_1 m_2, m_1 m_2+1$}}
 \State{$\hphantom{\texttt{N} \gets\ \ } $\scriptsize{$m_1 m_2+m_1, 2m_1 m_2]$}} \Comment{Initialize 25 indices in \texttt{N}, see Table~\ref{tab:hessian_indices}}
 \State{$\texttt{q} \gets 0$}

 \For{$\hat k \textup{ in } [0, m_z-1] $}
   \For{$\hat j \textup{ in } [0, m_y-1] $}
     \For{$\hat i \textup{ in } [0, m_x-1] $}
       \State{$i \gets \hat i + m_x \hat j + m_x m_y \hat k$}  \Comment{Compute linear index}
       \State{$i \pm x, i \pm y, i\pm z \gets \text{as in \eqref{eq:boundary_conditions}}$} \Comment{Compute neighbor indices}
       \State{$[\texttt{dTx},\texttt{dTy},\texttt{dTz}] \gets \texttt{imageDerivative}(T_{i+\texttt{N}})$} 
       \State{} \Comment{Compute image derivatives at 25 points}
       
         \For{$\texttt{k} \textup{ in } [0,24] $} \Comment{Compute 25 values of $dr^\top dr$, see \eqref{eq:hessMult}}
         \State{$\texttt{drdr}\gets 0$}
         \State{}
           \For{$(\tilde i, \tilde j) \textup{ in } N^{-1}(\texttt{N[k]}) $}
           \State{}\Comment{Compute 49 values of $\hat \rho_i$, see \eqref{eq:drTdr} and Table~\ref{tab:hessian_indices}}
             \State{$\texttt{drdr}\ \gets \texttt{drdr} + \hat \rho_{i+\tilde i}(-\tilde i) \cdot \hat \rho_{i+\texttt{N[k]}+\tilde j}(-\tilde j)$}
           \EndFor
           \State{}
           
           \State{$\texttt{q[$i$\hphantom{${}+2\bar m$}]} \gets \texttt{q[$i$\hphantom{${}+2\bar m$}]}+\ \texttt{dTx[12]} \cdot \texttt{drdr} \cdot \texttt{dTx[k]} \cdot \hat{\mathrm p}_{i+\texttt{N[k]}}$}
           \State{$\texttt{q[$i+\hphantom{2}\bar m $]} \gets \texttt{q[$i+\hphantom{2}\bar m $]}+\ \texttt{dTy[12]} \cdot \texttt{drdr} \cdot \texttt{dTy[k]} \cdot \hat{\mathrm p}_{i+\texttt{N[k]}+\bar m}$}
           \State{$\texttt{q[$i+2\bar m$]} \gets \texttt{q[$i+2\bar m$]}+\ \texttt{dTz[12]} \cdot \texttt{drdr} \cdot \texttt{dTz[k]} \cdot \hat{\mathrm p}_{i+\texttt{N[k]}+ 2\bar m}$}
           \State{} \Comment{The 12th element corresponds to the derivative at index $i$}

           \EndFor

     \EndFor
   \EndFor
 \EndFor
 \Ensure $\texttt{q[$i+d\bar m$]} = \hat{\mathrm q}_{d\bar m+i} $
\end{algorithmic}
\caption{Pseudocode for matrix-free computation of the elements of the result vector $\mathbf{\hat q}$ in the NGF Hessian-vector multiplication $\hat H \mathbf{\hat p} = \mathbf{\hat q}$ as in \eqref{eq:HNGF}. The algorithm can be fully parallelized over all loop iterations of $\hat i, \hat j, \hat k$, computing three elements of the result per thread.} \label{alg:hess_NGF}
\end{algorithm}

\subsection{Derivative computations for curvature regularization}

In \eqref{eq:curvatureMF}, a matrix-free version of the curvature gradient computation was already given. As the curvature computation is a quadratic function, the Hessian-vector multiplication
\begin{align}
  \mathrm q_{i+d\bar m^{\mathrm y}} 
  = \left(\nabla^2 S\mathbf p \right)_{i+d\bar m^{\mathrm y}} 
  = 2 \bar h^\mathrm y \tilde \Delta \left(\tilde \Delta  \mathbf p\right)_{i + d\bar m^{\mathrm y}}, \label{eq:curv_hessMult}
\end{align}
for $d \in \{0,1,2\}$, is closely related, with $\mathbf p$ replacing $\mathbf u$ in \eqref{eq:curvatureMF}. In contrast to NGF, this uses the exact Hessian rather than a Gauss-Newton approximation. As the regularizer operates on the deformation grid, no grid conversion is required.

\subsection{Grid conversion} \label{sec:grid-conversion}

We now consider the grid conversion steps omitted in the previous sections. The grid conversion maps deformation values $\mathbf{y} \in \mathbb R^{3\bar m^{\mathrm y}}$ from the deformation grid to the image grid using the function $P: \mathbb R^{3\bar m^{\mathrm y}} \to \mathbb R^{3\bar m}$ (Section~\ref{sec:discretization}). In this section, we separately analyze $P$ and its transpose $P^\top$. 

The conversion, i.e., interpolation, of $\mathbf{y}$ from deformation grid to image grid is performed using tri-linear interpolation and thus can be written as a matrix-vector product 
\begin{align*}
\mathbf{\hat y} = P \mathbf{y}\ \text{with}\ P \in \mathbb R^{3\bar m \times 3\bar m^{\mathrm y}}\ \text{and}\  \mathbf{y} \in \mathbb R^{3\bar m^{\mathrm y}},\ \mathbf{\hat y} \in \mathbb R^{3\bar m}.
\end{align*}
The transformation matrix $P$ has a block-diagonal structure with three identical blocks, each converting one of the three coordinate components of $\mathbf{y}$.
As the deformation grid is a nodal grid, no boundary handling is required, as each (cell-centered) image grid point is surrounded by deformation grid points.

In order to get closed-form expressions for the non-zero entries of $P$, define the function $c_l(k) := \frac{(k+0.5) m_l^{\mathrm y}}{m_l}$, which converts a coordinate index from deformation to image grid, and the remainder function $r_l(k) := c_l(k) - \lfloor c_l(k) \rfloor$. Then, for the point at coordinates $(\hat i, \hat j, \hat k)$ with $i= \hat i + \hat j m_x + \hat k m_x m_y$ and $d \in \{0,1,2\}$, the interpolated deformation value can be written as
\begin{align}
 \left(P \mathbf y\right)_{\hat i + \hat j m_x + \hat k m_x m_y + d\bar m } = \sum_{\alpha=0}^1 \sum_{\beta=0}^1 \sum_{\gamma=0}^1 w^{\alpha,\beta,\gamma}(\hat i, \hat j, \hat k)\ \mathrm y_{\xi(\alpha,\beta,\gamma,\hat i, \hat j, \hat k,d)},\label{eq:grid_conversion}
\end{align}
where the indices of the neighboring deformation grid points are
\begin{align}
 \xi(\alpha,\beta,\gamma,\hat i, \hat j, \hat k, d) := \left(\lfloor c_x(\hat i) \rfloor + \alpha\right)+ m_x\left( \lfloor c_y(\hat j) \rfloor + \beta\right) + m_x m_y\left( \lfloor c_z(\hat k) \rfloor + \gamma\right)+d\bar m \label{eq:xi}
 \end{align}
and the interpolation weights are given by
\begin{align*}
 w^{\alpha,\beta,\gamma}(\hat i, \hat j, \hat k) := \hat w^\alpha_x(\hat i)\hat w^\beta_y(\hat j) \hat w^\gamma_z(\hat k), \quad \hat w^s_l(k) := \begin{cases}1-r_l(k),\ &\text{if}\ s = 0,\\ r_l(k), \ &\text{otherwise.}\end{cases}
\end{align*}
As can be seen in \eqref{eq:grid_conversion}, a single interpolated deformation value on the image grid is simply a weighted sum of the values of its eight neighbors on the deformation grid, see also  Figure~\ref{fig:deformationToImageGrid}.

For the NGF derivative computations in \eqref{eq:NGF_discretized_gradient} and \eqref{eq:NGF_discretized_hessian}, the left-sided multiplication $\mathbf{\hat y}^\top P = (P^\top \mathbf{\hat y})^\top$ is also required in order to evaluate the transposed operator. For each (nodal) deformation grid point, this amounts to a weighted sum of values from all image grid points in the adjacent deformation grid cells, which can be many more than eight (Figure~\ref{fig:imageToDeformatioGridNaive}).

By singling out the contribution of all image grid points within a single deformation grid cell as in Figure~\ref{fig:imageToDeformatioGridRedBlack}, the weights 
$\hat w_l^s(k)$ can be re-used and only need to be computed once for each deformation grid cell. This suggests to parallelize the computations on a per-cell basis.
However, as a deformation grid point has multiple neighboring grid cells, this would lead to write conflicts. 

Therefore, in order to achieve both efficient re-use of weights as well as parallelizability, we propose to use a red-black scheme as shown in Figure~\ref{fig:imageToDeformatioGridRedBlack}. Each parallel unit sequentially processes a single  2D slice of deformation grid cells, first all odd (red) and finally all even slices (black).

\begin{figure}[tb]
\centering

\subfigure[]{
\begin{tikzpicture}[scale=0.51]

\draw [black, xstep=3.0cm,ystep=2.3cm, xshift=-0.5cm, yshift=-1cm, dashed] (0,0) grid (6,4.9);
\draw[dotted] (5.5cm,3.6cm) -- (6.5cm,3.6cm);
\draw[dotted] (5.5cm,1.3cm) -- (6.5cm,1.3cm);
\draw[dotted] (5.5cm,-1cm) -- (6.5cm,-1cm);

\draw[dotted] (-.5cm,-1cm) -- (-.5cm,-2cm);
\draw[dotted] (2.5cm,-1cm) -- (2.5cm,-2cm);
\draw[dotted] (5.5cm,-1cm) -- (5.5cm,-2cm);

\foreach \x in {0.3,0.9,1.5,2.1}
{   \foreach \y in {0.1,-0.5,-1.1,-1.7}
    {   \pgfmathtruncatemacro{\yPlot}{\y*-1}
        \node[rectangle,draw=black,fill=white,minimum width = 10, minimum height = 10, scale=0.5] at (\x*3cm-0.5cm,\y*-2cm-0.5cm) {};
    }
}

\foreach \x in {0,...,2}
{   \foreach \y in {0,...,-2}
    {   \pgfmathtruncatemacro{\yPlot}{\y*-1}
        \node[circle,draw=black,fill=white,scale=0.6] at (\x*3cm-0.5cm,\y*-2.3cm-1cm) {};
    }
}

\draw[->,shorten <=0.1cm,shorten >=0,>=stealth,semithick] (-0.5cm,3.6cm) -- (0.4cm,2.9cm);
\node[] at (-0.6cm,2.9cm){\small{$w^{0,0}$}};
\draw[->,shorten <=0.1cm,shorten >=0,>=stealth,semithick] (2.5cm,3.6cm) -- (0.4cm,2.9cm);
\node[] at (1.5cm,3.5cm){\small{$w^{1,0}$}};
\draw[->,shorten <=0.1cm,shorten >=0,>=stealth,semithick] (2.5cm,1.3cm) -- (0.4cm,2.9cm);
\node[] at (2cm,2.4cm){\small{$w^{1,1}$}};
\draw[->,shorten <=0.1cm,shorten >=0,>=stealth,semithick] (-0.5,1.3cm) -- (0.4cm,2.9cm);
\node[] at (-0.6cm,2.2cm){\small{$w^{0,1}$}};

\node[left=-0.5cm, above=0.02cm] at (0*3cm-0.5cm,-2*-2cm-0.5cm) {\small{$\mathbf y_{(0,0)}$}};
\node[right=0.01cm,above=0.02cm] at (1*3cm-0.5cm,-2*-2cm-0.5cm) {\small{$\mathbf y_{(1,0)}$}};
\node[left=-0.5cm, below=0.02cm] at (0*3cm-0.5cm,-2*-1cm-0.6cm) {\small{$\mathbf y_{(0,1)}$}};
\node[right=0.01cm,below=0.02cm] at (1*3cm-0.5cm,-2*-1cm-0.6cm) {\small{$\mathbf y_{(1,1)}$}};

\end{tikzpicture}\label{fig:deformationToImageGrid}  
}
\subfigure[]{
\begin{tikzpicture}[scale=0.51]

\draw [black, xstep=3.0cm,ystep=2.3cm, xshift=-0.5cm, yshift=-1cm, dashed] (0,0) grid (6,4.9);
\draw[dotted] (5.5cm,3.6cm) -- (6.5cm,3.6cm);
\draw[dotted] (5.5cm,1.3cm) -- (6.5cm,1.3cm);
\draw[dotted] (5.5cm,-1cm) -- (6.5cm,-1cm);

\draw[dotted] (-.5cm,-1cm) -- (-.5cm,-2cm);
\draw[dotted] (2.5cm,-1cm) -- (2.5cm,-2cm);
\draw[dotted] (5.5cm,-1cm) -- (5.5cm,-2cm);

\foreach \x in {0.3,0.9,1.5,2.1}
{   \foreach \y in {0.1,-0.5,-1.1,-1.7}
    {   \pgfmathtruncatemacro{\yPlot}{\y*-1}
        \node[rectangle,draw=black,fill=white,minimum width = 10, minimum height = 10, scale=0.5] at (\x*3cm-0.5cm,\y*-2cm-0.5cm) {};
    }
}

\foreach \x in {0,...,2}
{   \foreach \y in {0,...,-2}
    {   \pgfmathtruncatemacro{\yPlot}{\y*-1}
        \node[circle,draw=black,fill=white,scale=0.6] at (\x*3cm-0.5cm,\y*-2.3cm-1cm) {};
    }
}

\draw[->,shorten <=0.05cm,shorten >=0.1cm,>=stealth,semithick] (0.4cm,2.9cm) -- (2.5cm,1.3cm);
\draw[->,shorten <=0.05cm,shorten >=0.1cm,>=stealth,semithick] (2.2cm,2.9cm) -- (2.5cm,1.3cm);
\draw[->,shorten <=0.05cm,shorten >=0.1cm,>=stealth,semithick] (4cm,2.9cm) -- (2.5cm,1.3cm);

\draw[->,shorten <=0.05cm,shorten >=0.1cm,>=stealth,semithick] (0.4cm,1.7cm) -- (2.5cm,1.3cm);
\draw[->,shorten <=0.05cm,shorten >=0.1cm,>=stealth,semithick] (2.2cm,1.7cm) -- (2.5cm,1.3cm);
\draw[->,shorten <=0.05cm,shorten >=0.1cm,>=stealth,semithick] (4cm,1.7cm) -- (2.5cm,1.3cm);
                     
\draw[->,shorten <=0.05cm,shorten >=0.1cm,>=stealth,semithick] (0.4cm,0.5cm) -- (2.5cm,1.3cm);
\draw[->,shorten <=0.05cm,shorten >=0.1cm,>=stealth,semithick] (2.2cm,0.5cm) -- (2.5cm,1.3cm);
\draw[->,shorten <=0.05cm,shorten >=0.1cm,>=stealth,semithick] (4cm,0.5cm) -- (2.5cm,1.3cm);

\draw[->,shorten <=0.05cm,shorten >=0.1cm,>=stealth,semithick] (0.4cm,-0.7cm) -- (2.5cm,1.3cm);
\draw[->,shorten <=0.05cm,shorten >=0.1cm,>=stealth,semithick] (2.2cm,-0.7cm) -- (2.5cm,1.3cm);
\draw[->,shorten <=0.05cm,shorten >=0.1cm,>=stealth,semithick] (4cm,-0.7cm) -- (2.5cm,1.3cm);

\node at (4.1cm,0.9cm) {\textcolor{black}{\small{$\mathbf y_{(1,1)}$}}};

\end{tikzpicture}\label{fig:imageToDeformatioGridNaive}
}
\subfigure[]{
\begin{tikzpicture}[scale=0.51]

\draw [black, xstep=3.0cm,ystep=2.3cm, xshift=-0.5cm, yshift=-1cm, dashed] (0,0) grid (6,4.9);
\draw[dotted] (5.5cm,3.6cm) -- (6.5cm,3.6cm);
\draw[dotted] (5.5cm,1.3cm) -- (6.5cm,1.3cm);
\draw[dotted] (5.5cm,-1cm) -- (6.5cm,-1cm);

\draw[dotted] (-.5cm,-1cm) -- (-.5cm,-2cm);
\draw[dotted] (2.5cm,-1cm) -- (2.5cm,-2cm);
\draw[dotted] (5.5cm,-1cm) -- (5.5cm,-2cm);

\foreach \x in {0.3,0.9,1.5,2.1}
{  
   \foreach \y in {0.1,-0.5}
    {   \pgfmathtruncatemacro{\yPlot}{\y*-1}
        \node[rectangle,draw=black,fill=black,minimum width = 10, minimum height = 10, scale=0.5] at (\x*3cm-0.5cm,\y*-2cm-0.5cm) {};
    }
    
    \foreach \y in {-1.1,-1.7}
    {   \pgfmathtruncatemacro{\yPlot}{\y*-1}
        \node[rectangle,draw=black,fill=red,minimum width = 10, minimum height = 10, scale=0.5] at (\x*3cm-0.5cm,\y*-2cm-0.5cm) {};
    }
}

\foreach \x in {0,...,2}
{   \foreach \y in {0,...,-2}
    {   \pgfmathtruncatemacro{\yPlot}{\y*-1}
   
        \node[circle,draw=black,fill=white,scale=0.6] at (\x*3cm-0.5cm,\y*-2.3cm-1cm) {};
    }
}

\draw[->,shorten <=0.05cm,shorten >=0.1cm,>=stealth,semithick] (0.4cm,2.9cm) -- (-0.5cm,3.6cm);
\draw[->,shorten <=0.05cm,shorten >=0.1cm,>=stealth,semithick] (0.4cm,2.9cm) -- (2.5cm,3.6cm);
\draw[->,shorten <=0.05cm,shorten >=0.1cm,>=stealth,semithick] (0.4cm,2.9cm) -- (2.5cm,1.3cm);
\draw[->,shorten <=0.05cm,shorten >=0.1cm,>=stealth,semithick] (0.4cm,2.9cm) -- (-0.5,1.3cm);
                     
\draw[->,shorten <=0.05cm,shorten >=0.1cm,>=stealth,semithick] (2.2cm,2.9cm) -- (-0.5cm,3.6cm);
\draw[->,shorten <=0.05cm,shorten >=0.1cm,>=stealth,semithick] (2.2cm,2.9cm) -- (2.5cm,3.6cm);
\draw[->,shorten <=0.05cm,shorten >=0.1cm,>=stealth,semithick] (2.2cm,2.9cm) -- (2.5cm,1.3cm);
\draw[->,shorten <=0.05cm,shorten >=0.1cm,>=stealth,semithick] (2.2cm,2.9cm) -- (-0.5,1.3cm);
                     
\draw[->,shorten <=0.05cm,shorten >=0.1cm,>=stealth,semithick] (2.2cm,1.7cm) -- (-0.5cm,3.6cm);
\draw[->,shorten <=0.05cm,shorten >=0.1cm,>=stealth,semithick] (2.2cm,1.7cm) -- (2.5cm,3.6cm);
\draw[->,shorten <=0.05cm,shorten >=0.1cm,>=stealth,semithick] (2.2cm,1.7cm) -- (2.5cm,1.3cm);
\draw[->,shorten <=0.05cm,shorten >=0.1cm,>=stealth,semithick] (2.2cm,1.7cm) -- (-0.5,1.3cm);
                     
\draw[->,shorten <=0.05cm,shorten >=0.1cm,>=stealth,semithick] (0.4cm,1.7cm) -- (-0.5cm,3.6cm);
\draw[->,shorten <=0.05cm,shorten >=0.1cm,>=stealth,semithick] (0.4cm,1.7cm) -- (2.5cm,3.6cm);
\draw[->,shorten <=0.05cm,shorten >=0.1cm,>=stealth,semithick] (0.4cm,1.7cm) -- (2.5cm,1.3cm);
\draw[->,shorten <=0.05cm,shorten >=0.1cm,>=stealth,semithick] (0.4cm,1.7cm) -- (-0.5,1.3cm);

\node[left=-0.5cm, above=0.02cm] at (0*3cm-0.5cm,-2*-2cm-0.5cm) {\small{$\mathbf y_{(0,0)}$}};
\node[right=0.01cm,above=0.02cm] at (1*3cm-0.5cm,-2*-2cm-0.5cm) {\small{$\mathbf y_{(1,0)}$}};
\node[left=-0.5cm, below=0.02cm] at (0*3cm-0.5cm,-2*-1cm-0.6cm) {\small{$\mathbf y_{(0,1)}$}};
\node[right=0.01cm,below=0.02cm] at (1*3cm-0.5cm,-2*-1cm-0.6cm) {\small{$\mathbf y_{(1,1)}$}};

\end{tikzpicture} \label{fig:imageToDeformatioGridRedBlack}
}
\caption{Grid conversion operations (2D example). \textbf{Circles:} nodal deformation grid; \textbf{squares:} cell-centered image grid. \subref{fig:deformationToImageGrid} deformation grid to image grid: each value on the image grid is computed by a weighted sum of the values of its deformation grid neighbors. \subref{fig:imageToDeformatioGridNaive} transposed operation: each value on the deformation grid is a weighted sum of all image grid values from neighboring deformation grid cells.
\subref{fig:imageToDeformatioGridRedBlack} transposed operation using the proposed red-black scheme. Weighted values on the image grid are accumulated into all surrounding deformation grid points using multiple write accesses. Weights need to be computed only once per cell. Parallelization is performed per row (2D) or slice (3D) using a red-black scheme to avoid write conflicts.
}
\label{fig:red_black}
\end{figure}

\subsection{Extension to related models} \label{sec:extension_to_related_models}

By carefully examining the derivative structure of the NGF distance measure and curvature regularizer, we have derived closed-form expressions for computing individual elements of the gradients as well as Hessian matrix-vector products. The new formulations  operate directly on the input data, avoid storage of intermediate results (and thus costly memory accesses) and allow for fully parallel execution.

It is important to note that the proposed approach is by no means conceptually restricted to the NGF distance measure or curvature regularization. The central idea of replacing sparse matrix construction by on-the-fly coefficient calculation for derivative computation is in principle applicable to all distance measures and regularization schemes. The effectiveness, however, strongly depends on the derivative structure of the considered objective function terms. 

In more detail, consider an arbitrary distance measure $D = \psi(r(T(P)))$. We first observe that the proposed reformulation for the grid conversion operations related to $P$ as well as the image interpolation calculations for $T$ are independent of the choice of the distance metric. Hence, any matrix-free reformulation for these functions can be applied to every choice of distance measure $D$. Since the derivative of $\psi$ is just a vector, the most critical part for a matrix-free computation is the residual function derivative $\frac{\partial r}{\partial T}$. For NGF, the non-zero elements are distributed to just seven diagonals, thus allowing to derive a closed expression for $\frac{\partial r}{\partial T}$. In the case of the well-known sum-of-squared-differences (SSD) distance measure, the derivative $\frac{\partial r}{\partial T}$ is even the identity and can thus be omitted from the computations \cite{ruhaak2017matrix}.

More generally, matrix-free methods may be considered practically applicable as long as the sparsity pattern of $\frac{\partial r}{\partial T}$ is known and fixed. 
Examples of suitable matrix types are band matrices, block structures and similar configurations. Unfortunately, the widely used mutual information distance measure~\cite{collignon1995automated,viola1997alignment} does not allow for such application of the proposed matrix-free concept: here, the sparsity pattern of $\frac{\partial r}{\partial T}$ depends on the image intensity distribution of the deformed template image and thus on both $T$ and the current deformation $\vecy$ \cite{FAIR}. As the deformation changes at each iteration step, the sparsity pattern of $\frac{\partial r}{\partial T}$ may do so as well, prohibiting the derivation of static matrix-free calculation rules. Again, however, the matrix-free grid change schemes and the discussed template image derivative computations can directly be reused also for mutual information.

For the regularization term, we note that widely used energies such as the diffusive regularizer~\cite{fischer2002fast} or the linear-elastic potential \cite{modersitzki2004} are essentially calculated using finite difference schemes on the deformation $\vecy$. Hence, their matrix-free computation is far less complex than for the distance term with its function chain structure, and similar techniques may directly be used as done for the curvature regularization in this work. Also more complex regularization schemes such as hyperelastic regularization~\cite{burger2013hyperelastic}, which penalizes local volume change and generates diffeomorphic transformations, may be similarly implemented in a matrix-free manner. 

In summary, the underlying idea of our work is widely applicable and not restricted to the exemplary use case of NGF with curvature regularization.
An interesting question is what performance improvements can be expected for conceptually different methods such as LDDMM \cite{mang2017semi}; we refer to \cite{mang2016distributed} for some notes on parallelizing such methods.

\section{Algorithm analysis}\label{sec:algorithm_analysis}
In this section, we will perform a theoretical analysis and quantification of the expected speed-up obtained by using the proposed matrix-free approach, compared to a matrix-based approach as used in~\cite{FAIR}.

\subsection{Matrix element computations and memory stores}

While the exact cost of sparse matrix multiplications is hard to quantify and highly depends on the implementation and chosen sparse matrix format, meaningful estimates can be made for the cost of matrix coefficient computations and memory stores. As in this section we are primarily concerned with the \emph{time} consumed by memory write accesses, by ``memory store'' we refer to a single memory write operation. The \emph{amount} of memory required by each scheme is discussed in Section~\ref{sec:memory_requirements}.

\subsubsection{Matrix-based approach}
\paragraph{Data term}

For the NGF gradient, analyzing the structure of $\frac{\partial \psi}{\partial r}$, $\frac{\partial r}{\partial T}$, and $\frac{\partial T}{\partial P}$ in \eqref{eq:dPsi}, \eqref{eq:dr}, and \eqref{eq:dT}, an upper bound for the required number of non-zero element computations can be computed.

The vector $\frac{\partial \psi}{\partial r}$ requires the computation of $\bar m$ residual elements, where $\bar m$ is the number of cells in the image grid (Section \ref{sec:discretization}). For the matrix $\frac{\partial r}{\partial T}$, with a seven-banded structure, at most $7 \bar m$ elements need to be computed, while $\frac{\partial T}{\partial P}$ contains at most $3 \bar m$ non-zero matrix elements. 

The grid conversion, discussed in Section~\ref{sec:grid-conversion}, involves a matrix with eight coefficients in each row, one per contributing deformation grid point \eqref{eq:grid_conversion}, resulting in the computation of $8 \bar m$ coefficients. In total, this gives $19 \bar m$ coefficient computations for the NGF term in the matrix-based scheme. As the matrix elements only depend on the size of the image and deformation grids, they can be computed once and stored for each resolution.

\paragraph{Regularizer}
The curvature regularizer is a quadratic function and can be implemented using a matrix with $25$ diagonals, resulting in $25 \bar m^\text y$ memory stores, where $\bar m^\text y$ is the number of cells in the deformation grid. 

\paragraph{Hessian}
For the Gauss-Newton Hessian-vector multiplication, no additional coefficient computations are needed, as the Hessian is approximated from first-order derivatives.

\paragraph{Total number of operations}
After computing all matrix elements, these have to be stored in memory for later use, resulting in $19 \bar m + 25 \bar m^\text y$ memory store operations for one evaluation of the full gradient and Hessian approximation. This assumes that the latter will never be explicitly stored, as discussed in the following section. If multiple matrix-vector multiplications with the same gradient are required, as in the inner CG iterations, the factor matrices can be re-used and no new memory stores are necessary.

\subsubsection{Matrix-free approach}
\paragraph{Data term}
For the matrix-free formulation of the NGF gradient in~\eqref{eq:gradNGF}, excluding grid conversion, the same $11 \bar m$ coefficient computations as in the matrix-based case have to be performed for one evaluation of the gradient. 
However, no intermediate storage of elements is needed.

For the grid conversion, the coefficients are re-computed during each gradient evaluation and Hessian-vector multiplication. Furthermore, they are re-computed during the application of the transposed operator. With  $8 \bar m$ computations in each grid conversion, this adds up to $16 \bar m$ additional coefficient computations for application of the forward and transposed operator compared to the matrix-based case.

In the proposed approach, the grid conversions are evaluated separately from the remaining derivatives, instead of fully integrating them into the matrix-free computations. While the latter further lowers memory requirements, it requires re-computations of weights even within a single conversion, which in our experience resulted in slower overall computation times.

\paragraph{Regularizer}
For the curvature regularizer, the coefficients are highly redundant, resulting from the finite-differences stencil. Therefore, there is no computational cost for coefficients in the matrix-free scheme.

\paragraph{Hessian}
For the matrix-free Gauss-Newton Hessian-vector multiplication, a number of elements are re-computed. As shown in \eqref{eq:hessMult} and \eqref{eq:drTdr}, coefficients $\hat \rho_i(k)$ are computed $2 \cdot 49 \bar m $ times for one Hessian-vector multiplication. With 25 diagonals in $dr^\top dr$, image derivatives from $\frac{\partial T}{\partial P}$ are required for $25$ points, resulting in additional $25 \bar m$ coefficient computations. In total, the Hessian-vector multiplication requires $123 \bar m$ additional coefficient computations compared to the matrix-based method. 

Note that, in the matrix-based case, we have assumed that the full Hessian is never stored but rather evaluated using its factor matrices (Figure~\ref{fig:hess_mult_schema}). If the final Hessian is also saved, an additional $9 \cdot 25\bar m$ memory stores are required, as $dr^\top dr$ exhibits 25 diagonals.

While in both schemes the gradient is fully stored once computed and can be re-used, for the matrix-free Hessian-vector multiplication, all elements need to be re-computed for every single matrix-vector multiplication. Therefore, the exact number of additional arithmetic operations depends on other variables such as the number of CG iterations and line-search steps, and is specific to the input data. 

\paragraph{Total number of operations} In summary, for matrix-free NGF gradient evaluations, there are no additional coefficient computations compared to the matrix-based scheme and all intermediate memory stores are completely eliminated, making it highly amenable to massive parallelization.

For the Gauss-Newton Hessian-vector multiplications,  the matrix-free scheme adds a considerable number of re-computations, but again no stores are required. However, memory store operations are generally orders of magnitude slower than arithmetic operations. Therefore, as will be shown in the next section, in practice this trade-off between computations and memory stores  results in faster overall runtimes. Moreover, it significantly reduces the amount of required memory, which is a severely limiting factor for Hessian-based methods.

\subsection{Memory requirements} \label{sec:memory_requirements}

As discussed in the previous section, the memory required for the derivative matrices in a classical matrix-based scheme depends on the image grid size $\bar m$. For the gradient computations, when storing $\frac{\partial \psi}{\partial P}$, $\frac{\partial r}{\partial T}$ and $\frac{\partial T}{\partial P}$, $\bar m $, $7 \bar m$ and $3 \bar m$ values need to be saved. 
For the grid conversion, $8 \bar m$ values need to be stored, while the curvature regularizer requires $25 \bar m^\text y$ values. In terms of complexity,  this leads to memory requirements in the order of $\mathcal O(\bar m)$.

In comparison, the memory required by the matrix-free schemes does not depend on the image grid size. When integrating the grid conversion steps directly into the computation, as described in the previous section, only constant auxiliary space is required. The matrix-free approach therefore reduces the required memory size for intermediate storage
from $\mathcal O(\bar m)$ to $\mathcal O(1)$.

\section{Experiments and results}\label{sec:experiments}
In order to characterize the performance at the implementation level, we performed a range of experiments comparing four approaches for derivative computation. Three approaches are implemented on the CPU: an open-source, mixed C and MATLAB implementation in the FAIR toolbox~\cite{FAIR} ({``FAIR''}), an in-house C++ implementation of the matrix-based computations using sparse matrices ({``MBC''}) and a C++ implementation of the proposed matrix-free computations ({``MFC''}). Additionally, we evaluated a GPU implementation of the matrix-free computations using NVIDIA CUDA ({``MFC$^{\text{GPU}}$''}). 

All CPU implementations use OpenMP for parallelization of critical components. In {FAIR} and {MBC}, these include distance measure computations, linear interpolation and sparse-matrix multiplications. FAIR provides a ``matrix-free'' mode, which uses matrix-free computations for the curvature regularizer. In contrast to our approach, all other derivative matrices are still built explicitly and require additional storage and memory accesses.
This mode was used for all comparisons. Furthermore, MATLAB-MEX versions of FAIR components were used where available. In the matrix-free method {MFC}, full element-wise parallelization was used for function value and derivative calculations.

Both the classical {MBC} and proposed {MFC} implementations have undergone thorough optimization. In addition, due to the favorable structure, the NGF gradient computations for {MFC} on the CPU were manually vectorized using the Intel Advanced Vector Instructions (AVX). As stated in Section~\ref{sec:grid-conversion}, the grid conversions were computed as separate steps. Additionally, the deformed template was temporarily stored in order to reduce interpolation calls. 

The experiments on the CPU were performed 
on a dual processor 12-core Intel Xeon E5-2620 workstation with \SI{2.0}{\giga \hertz} and \SI{32}{\giga\byte} RAM. The GPU computations were performed on a NVIDIA GeForce GTX980, see Section~\ref{sec:gpu_implementation} for more detailed specifications.
All experiments were
averaged over 30 runs with identical input in order to reduce measurement noise.

\subsection{Scalability}\label{sec:ex_scalability}
The proposed matrix-free derivative calculations permit a fully parallel execution with only a single reduction for the function value, leading to high arithmetic intensity, i.e., floating point operations per byte of memory transfer. Thus, excellent scalability with respect to the number of computational cores can be expected.

In order to experimentally support this claim, we computed objective function derivatives using a varying number of computational cores on CPU. Two cases were considered: small images ($64^3$ voxels), typically occurring in multi-level computations of medical images, and larger images ($512^3$ voxels), representing, e.g., state-of-the-art thoracic computed tomography (CT) scans. 

As can be seen from the results in Table~\ref{table:scaling}, for both image sizes, all NGF evaluations -- function value, gradient, and Hessian-vector multiplication -- exhibit the expected speedup factors ranging from 11.0 to 15.0 on a 12-core system. Grid conversions from deformation grid to image grid also show speedups of approximately~12.  For the transposed operator, on small images, a speedup of only 4.34 is achieved: due to the small number of parallel tasks available, not all computational cores can be fully utilized. On the large images,  the expected speedup of 12.4 is achieved.

In comparison, the curvature regularizer scales much less favorably, with speedups ranging from 1.54 to 4.90. As the implementation of the curvature regularizer derivatives is fully parallelized and matrix-free, this indicates that the curvature computation is limited by memory bandwidth rather than arithmetic throughput. However, this has little effect on the overall performance, as the evaluation of the curvature regularizer is multiple orders of magnitude faster than the NGF evaluation.

\begin{table}
\centering
\begin{tabular}{rS[table-format=4.3, ,table-align-text-post = false]S[table-format=3.4]S[table-format=2.2]|S[table-format=3.4]S[table-format=2.4]S[table-format=2.2]}
&\multicolumn{3}{c}{Small images} & \multicolumn{3}{c}{Large images} \\
\text{Method} & \text{Serial} (\si{\milli\second})& \text{Par.} (\si{\milli\second})& \text{Speedup} & \text{Serial} (\si{\second})& \text{Par.} (\si{\second})& \text{Speedup}  \\
\toprule 
$D$  & 65.6 & 5.94 & \bfseries 11.0 & 44.6 & 3.31 & \bfseries 13.5 \\
$\frac{\partial D}{\partial P}$ & 131  & 10.6  & \bfseries 12.4 & 72.9 & 5.40  & \bfseries 13.5 \\
$P\mathbf y$ & 8.95 & 0.743  & \bfseries 12.1 & 4.51 & 0.355 & \bfseries 12.7\\
$P^\top \mathbf{\hat y}$ & 9.91  & 2.27 & 4.34 & 7.14& 0.58 & \bfseries 12.4 \\
$\hat H \mathbf{\hat p}$&1450 &117 & \bfseries 12.4 &913 &61.1 & \bfseries 15.0\\ \midrule
$S$&0.101&0.0655&1.54 &0.0553&0.0167 &3.31\\ 
$\nabla S$ &0.226&0.0411 &5.51 &0.110&0.0428 &2.58\\
$\nabla^2 S \mathbf{p}$ &0.224&0.0450 &4.98 &0.109 &0.0371 &2.94
\end{tabular}
\caption{Scaling of the proposed parallel implementation on a 12-core workstation compared to serial computation. Runtimes and speedups are itemized for gradient evaluations, Hessian vector-multiplications and grid conversions. All operations scale almost linearly, with the exception of the transposed grid conversion operator $P^\top \hat{\mathbf y}$ on small images, where the number of available parallel tasks is too low, and the curvature regularizer, which is primarily memory-bound. Small images: $64^3$ image resolution, $17^3$ deformation grid size; large images: $512^3$ image resolution, $129^3$ deformation grid size.}
\label{table:scaling}
\end{table}

\subsection{Runtime comparison} \label{sec:runtime_comparison}
As discussed in Section~\ref{sec:algorithm_analysis}, the matrix-free approach reduces the memory consumption and improves parallelization, however it introduces additional cost due to multiple re-calculations. Therefore, it is interesting to compare in detail the runtime of the matrix-free implementation {MFC} to the matrix-based implementations {MBC} and {FAIR}.

\subsubsection{Derivative and Hessian evaluation}\label{sec:runtime_deriComp}

\begin{table}
\centering
\begin{tabular}
{rrrS[round-mode=places, round-precision=2]S[round-mode=places, round-precision=2]S[round-mode=places, round-precision=2]|S[round-mode=places, round-precision=1] S[round-mode=places, round-precision=1] S[round-mode=places, round-precision=1]}
\multicolumn{3}{c}{}&\multicolumn{3}{c}{\text{Runtime}}&\multicolumn{3}{c}{\text{Speedup}}\\
&&&&&& \text{\footnotesize{MBC}} & \text{\footnotesize{\textbf{MFC}}}&\text{\footnotesize{\textbf{MFC}}}\\
&\multirow{2}{*}{$m_i$}&\multirow{2}{*}{$m_i^\mathrm{y}$} &\text{\multirow{2}{*}{FAIR (\si{\second})}} & \text{\multirow{2}{*}{MBC (\si{\second})}} & \text{\textbf{MFC} (\si{\second})} &\text{\footnotesize{vs.}}&\text{\footnotesize{vs.}}&\text{\footnotesize{vs.}}\\
&&&&&\text{\footnotesize{(Proposed)}}&\text{\footnotesize{FAIR}}&\text{\footnotesize{FAIR}}&\text{\footnotesize{MBC}}\\
\toprule
\multirow{12}{*}{\rotatebox[origin=c]{90}{\enspace Gradient}}&
512&513&\text{*}    &\text{*}& 15.3896   & \text{--}    & \text{--}  & \text{--}  \\
&&129   &\text{*}    &\text{*}& 4.61269    & \text{--}    & \text{--}  & \text{--}  \\
&&33    &\text{*}    &\text{*}& 4.47692    & \text{--}    & \text{--}  & \text{--}  \\
\cmidrule{2-9}                  
&256&257&\text{*}        &\text{*}& 1.36904    & \text{--}    &  \text{--}   & \text{--}  \\              
&&65    &155.623487 &5.41759   & 0.565826   &28.7255932989    &\bfseries 275.0377094725    &\bfseries 9.5746572268   \\           
&&17    &129.903032 &5.36889   & 0.604857   &24.1955100589 &\bfseries 214.7665183672  &\bfseries 8.8762963808   \\         
\cmidrule{2-9}          
&128&129&19.565225  &1.69946   & 0.187177   &11.51261283 &\bfseries 104.5279334534 &\bfseries 9.0794274938\\               
&   &33 &17.250636  &0.781679  & 0.0683456  &22.0686957178 &\bfseries 252.4030222867  &\bfseries 11.4371517698\\           
&   &9  &14.445142  &0.799346  & 0.115058   &18.0712007066 &\bfseries 125.5466112743  &\bfseries 6.9473309114\\              
\cmidrule{2-9}            
&64&65  &2.305313   &0.260584  & 0.0514679 &8.8467173733 &\bfseries 44.7912776702 &\bfseries 5.0630392925\\                
&&17    &1.9353     &0.109551  & 0.0098314 &17.6657447216 &\bfseries 196.8488719816  &\bfseries 11.1429704823\\             
&&5     &1.578713   & 0.109776 & 0.0468285 &14.3812217607 &\bfseries 33.7126536191 &\bfseries 2.3442134598\\    
\cmidrule[0.75pt]{1-9}
\multirow{12}{*}{\rotatebox[origin=c]{90}{\enspace Hessian-vector mult.}}
&512 &513  &\text{*}  &\text{*}                   & 83.1433    &\text{--}  &\text{--}  & \text{--}   \\
&    &129  &\text{*}  &\text{*}                   & 64.4637     &\text{--}  &\text{--}  & \text{--}   \\
&    &33   &\text{*}  &\text{*}                   & 64.2767     &\text{--}  &\text{--}  & \text{--}   \\
\cmidrule{2-9}                                                 
&256 &257  &\text{*}  &\text{*}                   & 8.18287      &\text{--}  &\text{--}  & \text{--}   \\
&    &65   &\text{*}  &\text{*}                   & 7.51372      &\text{--}  &\text{--}  & \text{--}   \\
&    &17   &230.751899&\text{*}                  & 7.58328      &\text{--}  &\bfseries 30.4290358526  &\text{--}\\
\cmidrule{2-9}                                             
&128 &129  &\text{*}  &\text{*}                   & 1.06301      &\text{--}  &\text{--}  &\text{--}  \\
&    &33   &36.277982 &27.8366613            & 1.01057   &1.303244725& \bfseries 35.8985344904&\bfseries 27.5455053089\\
&    &9    &24.441982 &21.3745204            & 1.01479   &1.1435101954&\bfseries 24.0857537027&\bfseries 21.06299865\\
\cmidrule{2-9}                                             
&64 &65    &25.196309 &20.54845              & 0.128854  &1.2261902479&\bfseries 195.5415353811 &\bfseries 159.4707964052\\
&   &17    &4.050656  &3.38442194            & 0.124799  &1.1968531323&\bfseries 32.4574395628&\bfseries 27.1189828444\\
&   &5     &2.561991  &2.519195458           & 0.140871  &1.0169877815&\bfseries 18.1867879123&\bfseries 17.8829954923\\
\end{tabular}
\caption{Runtimes (left) and speedup factors (right) for objective function gradient computations (top) and Hessian matrix-vector multiplications (bottom) on CPU. Three implementations were compared: a classical matrix-based implementation from the FAIR toolbox (FAIR), an optimized matrix-based C++ implementation (MBC), and the proposed matrix-free approach (MFC). For each image grid size~$m_i$, three different deformation grid sizes $m_i^\mathrm{y}$ were tested.
The proposed matrix-free approach {MFC} achieves speedups of $2.34$--$11.4$ (derivatives) and $17.9$--$159$ (Hessian multiplications) compared to the optimized C++ implementation MBC (rightmost column). It can also handle high resolutions for which FAIR and MFC run out of memory (third and fourth column, marked $*$).}
\label{table:objfunGradRuntime}
\end{table}

We separately evaluated runtimes for computation of the derivatives and for matrix-vector multiplication with the Gauss-Newton Hessian, including grid conversions.%

Image sizes ranged from $64^3$ to $512^3$ voxels. To account for various use cases with different requirements for accuracy, we tested three different resolutions for the deformation grid: full resolution, $1/4$, and $1/16$ of the image resolution. 
As can be seen from the results in Table~\ref{table:objfunGradRuntime}, for the gradient computation, we achieve speedups from 33.7 to 275 relative to {FAIR}. Furthermore, {FAIR} runs out of memory at larger image and deformation grid sizes. Here, the matrix-free scheme allows for computations at much higher resolutions.

Relative to the matrix-based C++ implementation {MBC}, we achieve speedups between 2.34 and 11.4. These values are particularly remarkable if one considers that {MBC} is already highly optimized and parallelized, and that {MFC} includes a large amount of redundant computations. Again, {MFC} allows to handle the highest resolution of $512^3$, while {MBC} runs out of memory at $256^3$ voxels. 

Comparing the runtimes of the Hessian-vector multiplication in Table~\ref{table:objfunGradRuntime}, the benefits of the low memory usage of {MFC} become even more obvious. While {MBC} and {FAIR} cannot be used for more than half of the resolutions tested, {MFC} has no such issues and scales approximately linear in the number of voxels.
MFC achieves particularly high speedups for large deformation grids compared to the optimized C++ implementation MBC, ranging from 17.9 to 159.

\subsubsection{Full registration}\label{sec:full_registration}

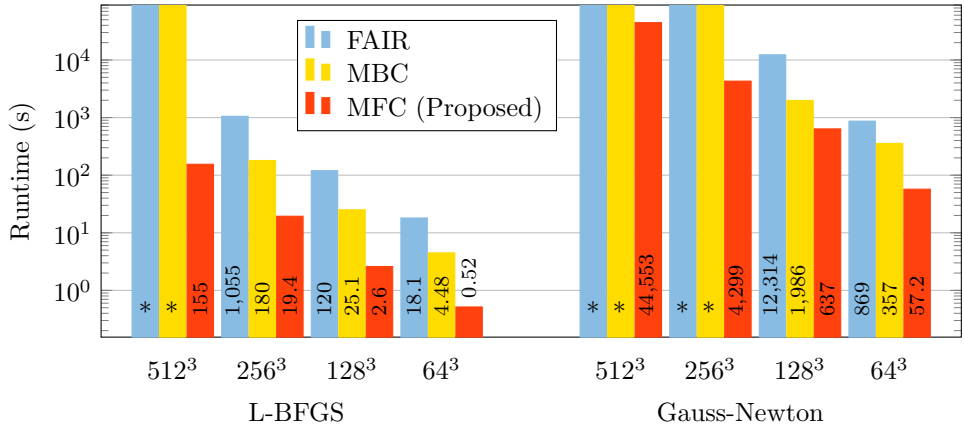
\begin{figure}[tb]
\begin{tikzpicture}
\begin{axis}[
        width  = \textwidth,
        height = 6cm,
        major x tick style = transparent,
        ybar=1*\pgflinewidth,
        bar width=10pt,
        ymajorgrids = true,
        ylabel = {Runtime (\si{\second})},
        symbolic x coords={512bfgs,256bfgs,128bfgs,64bfgs,empty,512gn,256gn,128gn,64gn},
        xtick = data,
        xticklabels = {$512^3$,$256^3$,$128^3$,$64^3$,$512^3$,$256^3$,$128^3$,$64^3$},
        scaled y ticks = false,
        enlarge x limits=0.1,
        ymode=log,
        ymax=90000,
        log origin=infty,
nodes near coords bottom/.style={
    scatter/position=absolute,
    close to zero/.style={
        at={(axis cs:\pgfkeysvalueof{/data point/x},\pgfkeysvalueof{/data point/y})},
        color=black,
    },
    big value/.style={
        at={(axis cs:\pgfkeysvalueof{/data point/x},0.25)},
        color=black, text opacity=1, 
        inner ysep=0.5pt,
    },
     every node near coord/.append style={
      check for zero/.code={%
        \pgfmathfloatifflags{\pgfplotspointmeta}{0}{%
            \pgfkeys{/tikz/coordinate}%
        }{%
            \begingroup
            \pgfkeys{/pgf/fpu}%
            \pgfmathparse{\pgfplotspointmeta<#1}%
            \global\let\result=\pgfmathresult
            \endgroup
            %
            %
            \pgfmathfloatcreate{1}{1.0}{0}%
            \let\ONE=\pgfmathresult
            \ifx\result\ONE
                \pgfkeysalso{/pgfplots/close to zero}%
            \else
                \pgfkeysalso{/pgfplots/big value}%
            \fi
        }
      },
      check for zero, 
      font=\footnotesize, 
      rotate=90, anchor=west, 
    },%
},%
nodes near coords={\pgfmathprintnumber{\pgfplotspointmeta}},
nodes near coords bottom=1,
        point meta=rawy,
        legend cell align=left,
        legend style={
                at={(0.53,0.62)},
                anchor=south east,
                column sep=1ex,
        }
    ]
        \addplot[style={FhCD02,fill=FhCD02,mark=none}]
            coordinates {(512bfgs, 10000000) (256bfgs,1055) (128bfgs,120.0) (64bfgs,18.1) (512gn, 10000000) (256gn,10000000) (128gn,12314) (64gn,869)};

        \addplot[style={FhCD19,fill=FhCD19,mark=none}]
             coordinates {(512bfgs,10000000) (256bfgs,180) (128bfgs,25.1) (64bfgs,4.48) (512gn, 10000000) (256gn,10000000) (128gn,1986) (64gn,357)};

        \addplot[style={redlk,fill=redlk,mark=none}]
             coordinates {(512bfgs,155) (256bfgs,19.4) (128bfgs,2.60) (64bfgs,0.516) (512gn, 44553) (256gn,4299) (128gn,637) (64gn,57.2)};

        \legend{FAIR,MBC,MFC (Proposed)}

    \end{axis}
    \node (a) at (0.59,0.35) {\textcolor{black}{*}};
    \node (b) at (0.59+0.35,0.35) {\textcolor{black}{*}};
    
    \node (c) at (0.59+5*1.19,0.35) {\textcolor{black}{*}};
    \node (c) at (0.59+5*1.19+0.35,0.35) {\textcolor{black}{*}};
    
    \node (c) at (0.59+6*1.19,0.35) {\textcolor{black}{*}};
    \node (c) at (0.59+6*1.19+0.35,0.35) {\textcolor{black}{*}};
    
    \node (c) at (2.6,-1.0) {L-BFGS};
    \node (c) at (8.5,-1.0) {Gauss-Newton};
\end{tikzpicture} 
\caption{Runtime comparison for a full multi-level registration on CPU using L-BFGS and Gauss-Newton optimization schemes for different image sizes (logarithmic scale). 
{FAIR}: matrix-based, MATLAB, using the FAIR toolbox \cite{FAIR}; MBC: matrix-based, C++, using sparse matrices; MFC: proposed matrix-free algorithm in C++; $*$: computation ran out of memory. Deformation grid sizes were set to one quarter of the image size in each dimension. 
The proposed algorithm (MFC) enables faster computations at higher resolutions for both optimization schemes and routinely handles resolutions of up to $512^3$ voxels.} \label{fig:multiLevelRuntime}
\end{figure}

The ultimate goal of the presented approach is to improve runtime and memory usage of the full image registration system in real-world applications. Therefore we performed a range of  full multi-level image registrations on three levels for different image and deformation grid sizes with all three implementations. 
The input data consisted of thorax-abdomen images as described in Section~\ref{sec:oncological_followup} below, cropped to a size of $512^3$ voxels. In order to avoid potentially large runtime differences caused by the effect of small numerical differences on the stopping criteria, we set a fixed number of $20$ iterations on each resolution level.

\begin{table}[tb]
\centering
\begin{tabular}{rS[table-format=1.0]S[table-format=2.2]S[table-format=2.2]S[table-format=2.2]|S[table-format=1.0]S[table-format=1.0]S[table-format=2.2]S[table-format=2.2]}
&\multicolumn{4}{c}{L-BFGS}&\multicolumn{4}{c}{Gauss-Newton}\\
&\text{$512^3$}&\text{$256^3$}&\text{$128^3$}&\text{$64^3$}&\text{$512^3$}&\text{$256^3$}&\text{$128^3$}&\text{$64^3$}\\
\midrule
\textbf{MFC} vs. FAIR &\text{*}&54.4&46.1&34.8&\text{*}&\text{*}&19.4&15.2\\
\textbf{MFC} vs. MBC  &\text{*}&9.27&9.67&8.62&\text{*}&\text{*}&3.12&6.25\\
\end{tabular}
\caption{Speedup factors of the proposed matrix-free approach on CPU (MFC is $n$ times faster) relative to the matrix-based MBC and MATLAB-based FAIR methods for the data in Figure \ref{fig:multiLevelRuntime}. $*$: FAIR or MBC ran out of memory.} \label{table:multiLevelSpeedup}
\end{table}

\begin{table}[t]
\centering
\begin{tabular}{rS[table-format=4.0]S[table-format=5.0]S[table-format=4.0]S[table-format=3.0]|S[table-format=4.0]S[table-format=3.0]S[table-format=4.0]S[table-format=3.0]}
&\multicolumn{4}{c}{L-BFGS}&\multicolumn{4}{c}{Gauss-Newton}\\
&\text{$512^3$}&\text{$256^3$}&\text{$128^3$}&\text{$64^3$}&\text{$512^3$}&\text{$256^3$}&\text{$128^3$}&\text{$64^3$}\\
\midrule
\textbf{MFC} &\bfseries 4914&\bfseries 618&\bfseries 78&\bfseries 10&\bfseries 4296&\bfseries 539&\bfseries 68&\bfseries 9\\
MBC          &\text{*}&14673&1837&230&\text{*}&\text{*}&5658&725\\
FAIR         &\text{*}&17422&2403&543&\text{*}&\text{*}&4375&800\\
\end{tabular}
\caption{Total peak memory usage in megabytes for the full multi-level registration on CPU, corresponding to the data in Figure~\ref{fig:multiLevelRuntime}. Reported sizes include objective function derivatives, multi-level representations of the images, memory required for the optimization algorithm, CG solver, and further auxiliary space. Compared to MBC and FAIR, the proposed algorithm (MFC) enables computations at high resolutions with moderate memory use. $*$: FAIR or MBC ran out of memory (more than \SI{32}{\giga\byte} required).} \label{table:multiLevelMemory}
\end{table}

For L-BFGS, the proposed method {MFC} is between 8.62 and 9.27 times faster than {MBC} (Figure~\ref{fig:multiLevelRuntime} and Table~\ref{table:multiLevelSpeedup}). Computation on the highest resolution with an image size of $512^3$ and deformation grid size of $129^3$ is only possible using {MFC} due to memory limitations of the other methods.

For Gauss-Newton, overall speedups range from 3.12 to 6.25. Additionally, due to the more memory-intense Hessian computations, even a registration with an image size of $256^3$ and a modest deformation grid size of $65^3$ could only be performed with the proposed {MFC} method.

While for L-BFGS the overall speedups are in a similar range as the raw speedups in Table~\ref{table:objfunGradRuntime}, they are slightly reduced for the Gauss-Newton method. This is due to the fact that in the matrix-based MBC method, the stored NGF Hessian is re-used multiple times in the CG solver, while it is re-computed for each evaluation in the matrix-free MFC approach. 
Solving a linear system with the CG solver using the NGF Hessian is only needed for Gauss-Newton optimization, which could additionally impose a performance bottleneck. In our evaluations, however, we found that, while the CG solver consumes $97\%$ to $99\%$ of the total registration runtime, the matrix-free NGF Hessian-vector multiplications again consume more than $99\%$ of the overall CG solver runtime, which can thus fully benefit from our parallelized computations.

The memory usage of the full multi-level registration for all approaches is shown in Table~\ref{table:multiLevelMemory}. Besides the objective function derivatives, the measured memory usage includes memory required for multi-level representations of the images, the optimization algorithm, CG solver and further auxiliary space. The matrix-free approach reduces memory requirements by one to two orders of magnitude. .
As the L-BFGS method uses additional buffers for storing previously computed gradients, the Gauss-Newton method has a lower memory usage with the matrix-free approach.

Overall, the proposed efficient matrix-free method enables both computation of higher resolutions and shorter runtimes.

\subsection{GPU Implementation} \label{sec:gpu_implementation} 

The presented matrix-free algorithm is not limited to a specific platform or processor type. Due to its parallelizable formulation and low memory requirements, it is also well-suited for implementation on specialized hardware such as GPUs. In comparison to CPUs, GPUs feature a massively parallel architecture and excellent computational performance. 
Thus, GPUs have become increasingly popular for general purpose computing applications \cite{shams2010survey}. However, in order to utilize these benefits, specialized implementations are required, which exploit platform-specific features such as GPU topology and different memory classes. 

In this section, we present a GPU implementation of the matrix-free registration algorithm using NVIDIA CUDA C/C++ \cite{nvidia2017cuda}. The CUDA framework allows for a comparatively easy implementation and direct access to hardware-related features such as different memory and caching models and has been widely adopted by the scientific community \cite{shams2010survey}.
We implemented optimized versions of all parts of the registration algorithm in CUDA, allowing the full registration algorithm to run on the GPU without intermediate transfers to the host. 
In the following, we present details on how the implementation makes use of the specialized platform features in order to improve performance.

\subsubsection{Implementation details}

The CUDA programming model organizes the GPU code in \emph{kernels}, which are launched from the host to execute on the GPU. The kernels are executed in parallel in different \emph{threads}, which are grouped to \emph{thread blocks}. 

\paragraph{Memory areas}

CUDA threads have access to different memory areas: Global memory, constant memory and shared memory. All data has to be transferred from the CPU main memory to the GPU global memory before it can be used for GPU computations. As frequent transfers can limit computational performance, the GPU implementation transfers the initial images $R,T$ exactly once at the beginning of the registration. All computations, including the creation of coarser images for the multi-level scheme, are then performed entirely in GPU memory.

While the global memory is comparatively large (up to \SI{16}{\giga\byte} on recent Tesla devices) and can be accessed from all threads, access is slow because caching is limited. Constant and shared memory in contrast are very fast. However, constant memory is read-only, while shared memory can only be accessed within the same thread block. Both are currently limited to 48 to 96 \si{\kilo\byte}, depending on the architecture.

We therefore store frequently used parameters, such as $m, m^{\mathrm y}, \bar m, h, h^{\mathrm y}, \bar h$ in constant memory to minimize access times. Shared memory (private per block) is used for reductions in L-BFGS and function value, with an efficient scheme from \cite{harris2007optimizing}.

\paragraph{Image interpolation}

Current GPUs include an additionally specialized memory area for textures. Texture memory stores the data optimized for localized access in 2D or 3D coordinates, and provides hardware-based interpolation and boundary condition handling. Therefore, we initially employed texture memory for the image interpolation of the deformed template image $T(\mathbf{\hat y})$. However, we found that the hardware interpolation causes errors in the order of $10^{-3}$ in comparison to the CPU implementation, as the GPU only uses 9-bit fixed-point arithmetic for these computations. The errors resulted in erroneous derivative calculations, causing the optimization to fail. Despite an interpolation speedup of approx. $35\%$ with texture memory, we therefore performed the interpolation in software within the kernels rather than in hardware.

\paragraph{Grid conversion}

In Section~\ref{sec:grid-conversion}, a red-black scheme for the transposed operator $P^\top \mathbf y$ was proposed in order to avoid write conflicts in parallel execution. While on the CPU this results in fast calculations, on the GPU there are specialized \emph{atomics} available to allow for fast simultaneous write accesses without conflicts. This enables a per-point parallel implementation without the red-black scheme, which also improves the limited speedup for small images shown in Table~\ref{table:scaling}. We took care that all GPU computations return identical results to the CPU code with the exception of numerical and rounding errors, in order to allow for a meaningful comparison.

\begin{figure}[t]
\centering
\subfigure[Reference]{\includegraphics[width=0.24\textwidth]{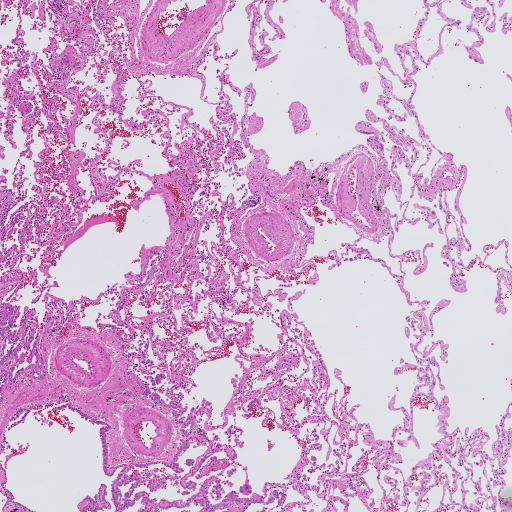}}
\subfigure[Template]{\includegraphics[width=0.24\textwidth]{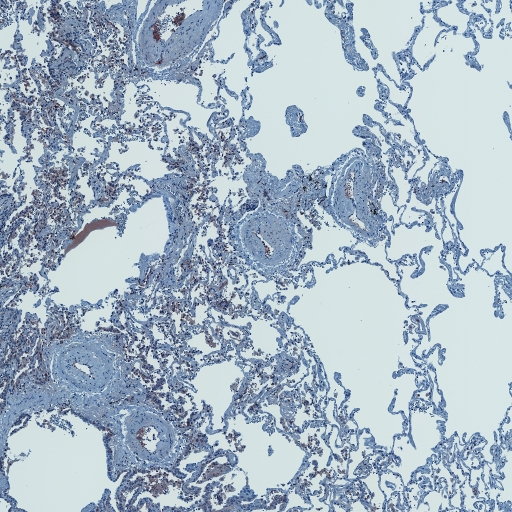}}
\subfigure[Initial overlay]{\includegraphics[width=0.24\textwidth]{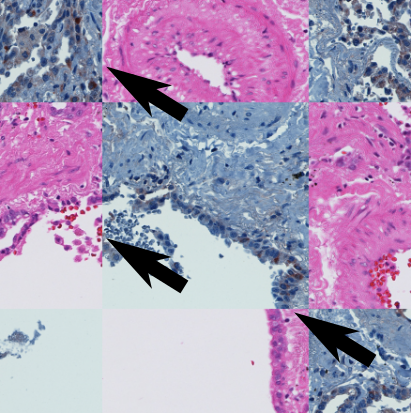}}
\subfigure[After registration]{\includegraphics[width=0.24\textwidth]{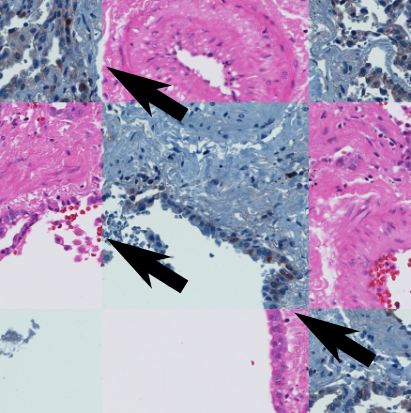}}
\caption{Registration of patches from histological serial sections with different stains on GPU. (a) reference image, (b) template image, (c) checkerboard overlay of details from both images before, and (d) after registration. While discontinuities can be seen at the borders of the checker pattern before registration, smooth transitions are achieved after registration. Areas of large differences are indicated by arrows. The images were obtained from the dataset used in \cite{lotz2016patch}.
}\label{fig:histo}
\end{figure}

\begin{table}[tb]
\centering
\begin{tabular}{rrS[round-mode=figures, round-precision=3]S[round-mode=figures, round-precision=3]|S[round-mode=figures, round-precision=3]}
\multicolumn{2}{c}{}&\multicolumn{2}{c}{\text{Runtime}}&\multicolumn{1}{c}{\text{Speedup}}\\
\text{$m_i$} & \text{$m_i^\mathrm{y}$} & \text{MFC (\si{\milli\second})} & \text{\textbf{MFC$^{\text{GPU}}$} (\si{\milli\second})} & \text{\footnotesize{\textbf{MFC$^{\text{GPU}}$} vs. MFC}}\\
\toprule
4096&2049   &0357       &0035.3 & \bfseries 10.1    \\
    &1025   &0289       &0033.5 & \bfseries 8.63    \\
    &513    &0286       &0032.3 & \bfseries 8.85    \\
\cmidrule{1-5}             
2048&1025   &0096.7     &0010.0 & \bfseries 9.67  \\              
    &513    &0072.3     &0009.38& \bfseries 7.71  \\           
    &257    &0074.3     &0009.32& \bfseries 7.97  \\         
\cmidrule{1-5}               
1024&513    &0019.7     &0003.66& \bfseries 5.38  \\              
    &257    &0016.2     &0003.53& \bfseries 4.59  \\           
    &129    &0016.0     &0002.93& \bfseries 5.46  \\         
\cmidrule{1-5}               
512&257     &0004.67    &0002.29& \bfseries 2.04  \\              
    &129    &0004.02    &0001.67& \bfseries 2.41  \\           
    &65     &0003.96    &0001.67& \bfseries 2.37  \\         
\cmidrule{1-5}               
256&129     &0001.27    &0001.04& \bfseries 1.22  \\               
   &65      &0001.16    &0001.03& \bfseries 1.13  \\           
   &33      &0002.68    &0001.07& \bfseries 2.50  \\              
\end{tabular}
\caption{Runtimes (left) and speedup factors (right) for objective function gradient computations using the proposed matrix-free method, implemented on CPU (MFC) and GPU (MFC$^\text{GPU}$) in 2D. For each image grid size~$m_i$, three different deformation grid sizes $m_i^\mathrm{y}$ were tested. The GPU-based implementation achieves additional speedup factors of up to $10.1$ in comparison to the matrix-free CPU implementation. The speedup increases with larger image and deformation grid sizes.
}
\label{table:objfunGradRuntime_CUDA}
\end{table}

\subsubsection{Evaluation}

\begin{figure}[tb]
\subfigure[]{
\begin{tikzpicture}
\begin{axis}[
        width  = 7cm,
        height = 6cm,
        major x tick style = transparent,
        ybar=1*\pgflinewidth,
        bar width=10pt,
        ymajorgrids = true,
        ylabel = {Runtime (\si{\second})},
        symbolic x coords={4096bfgs, 2048bfgs, 1024bfgs, 512bfgs, 256bfgs},
        xtick = data,
        xticklabels = {$4096^2$,$2048^2$,$1024^2$,$512^2$,$256^2$},
        scaled y ticks = false,
        enlarge x limits=0.2,
        ymode=log,
        ymax=20,
        log origin=infty,
nodes near coords bottom/.style={
    scatter/position=absolute,
    close to zero/.style={
        at={(axis cs:\pgfkeysvalueof{/data point/x},\pgfkeysvalueof{/data point/y})},
        color=black,
    },
    big value/.style={
        at={(axis cs:\pgfkeysvalueof{/data point/x},0.07)},
        color=black, text opacity=1, 
        inner ysep=0.5pt,
    },
     every node near coord/.append style={
      check for zero/.code={%
        \pgfmathfloatifflags{\pgfplotspointmeta}{0}{%
            \pgfkeys{/tikz/coordinate}%
        }{%
            \begingroup
            \pgfkeys{/pgf/fpu}%
            \pgfmathparse{\pgfplotspointmeta+0.75<#1}%
            \global\let\result=\pgfmathresult
            \endgroup
            %
            %
            \pgfmathfloatcreate{1}{1.0}{0}%
            \let\ONE=\pgfmathresult
            \ifx\result\ONE
                \pgfkeysalso{/pgfplots/close to zero}%
            \else
                \pgfkeysalso{/pgfplots/big value}%
            \fi
        }
      },
      check for zero, 
      font=\footnotesize, 
      rotate=90, anchor=west, 
    },%
},%
nodes near coords={\pgfmathprintnumber[fixed,precision=4]{\pgfplotspointmeta}},
nodes near coords bottom=1,
        point meta=rawy,
        legend cell align=left,
        legend style={
                at={(0.9,0.7)},
                anchor=south east,
                column sep=1ex,
        }
    ]
        \addplot[style={redlk,fill=redlk,mark=none}]
            coordinates {(4096bfgs, 13.9) (2048bfgs, 3.35) (1024bfgs,0.853) (512bfgs, 0.238) (256bfgs,0.0865)};
            
        \addplot[style={red!20!white,fill=red!20!white,mark=none}]
            coordinates {(4096bfgs, 1.51) (2048bfgs, 0.560) (1024bfgs,0.267) (512bfgs, 0.143) (256bfgs,0.0833)};

        \legend{MFC,MFC$^{\text{GPU}}$}

    \end{axis}
\end{tikzpicture} 
\hspace*{0.25cm}
}
\subfigure[]{
\raisebox{30mm}{
\begin{tabular}{rS[table-format=2.2]}
& \multicolumn{1}{c}{\text{Speedup}}\\
\rule{0pt}{11pt}\text{$m_i$}&\text{\footnotesize{\textbf{MFC$^{\text{GPU}}$} vs. MFC}}\\
\toprule
\text{4096} & 9.21 \\
\text{2048} & 5.98 \\
\text{1024} & 3.19 \\
\text{512}  & 1.66 \\
\text{256}  & 1.04 \\
\end{tabular}
}
}
\caption{Runtime comparison for a full multi-level registration on CPU and GPU. (a) Runtimes for CPU-based (MFC) and GPU-based (MFC$^{\text{GPU}}$) implementations of the proposed matrix-free algorithm for different image sizes (logarithmic scale), (b) corresponding speedup factors. Deformation grid sizes were set to one quarter of the image size in each dimension. In comparison to the CPU implementation, the GPU implementation achieves an additional speedup of up to $9.21$. The largest runtime improvements are achieved for high image resolutions.
}\label{table:multiLevelRuntime_CUDA}
\end{figure}
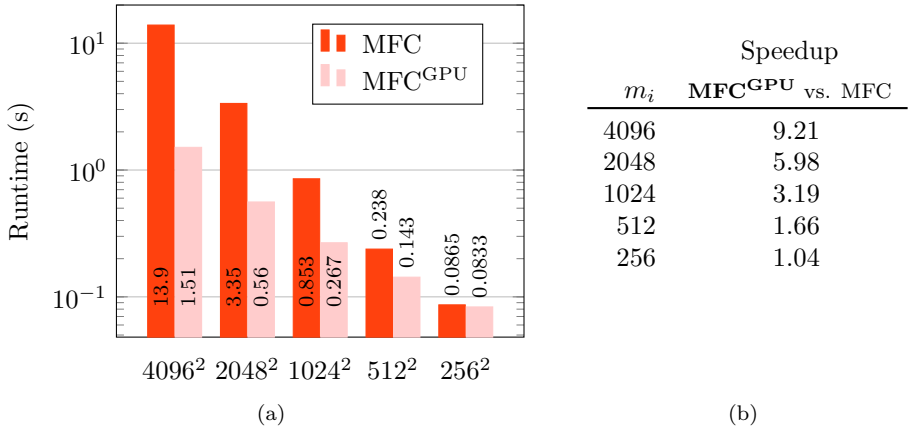

Experiments were performed on a GeForce GTX980 graphics card with \SI{4}{\giga\byte}
 of memory. The GPU features $16$ streaming multiprocessors with $128$ CUDA cores each, resulting in $2048$ CUDA cores in total for parallelization,
 and a theoretical peak performance of \SI{4.6}{\tera FLOPs} for single-precision computations. The double-precision peak performance is much lower at \SI{144}{\giga FLOPs}, which is less than the \SI{192}{\giga FLOPs} achieved by the dual-CPU Xeon E5-2620 that was used for the CPU computations. Additionally, the previously mentioned \textit{atomics} are only available for the single-precision \emph{float} datatype. Therefore all computations were performed in single precision. In order to minimize the effect of rouding errors, 
 an exception was made for reductions, which were computed in double precision.
 
 Comparing the peak performances of CPU and GPU, the theoretical maximum speedup is a factor of 20. While this maximum value is hard to obtain in practice, we will see that using the GPU can still lead to a substantial acceleration of the registration algorithm. To this end, we implemented a 2D image registration with the L-BFGS optimizer. As GPU implementations of FAIR and MBC are not available, we compare the results to the matrix-free CPU implementation MFC.
 
\paragraph{Derivative}

Similar to Section~\ref{sec:runtime_deriComp} and Table~\ref{table:objfunGradRuntime}, we measured the runtime for the evaluation of the objective function derivative. The results are compared with the matrix-free algorithm on CPU in Table~\ref{table:objfunGradRuntime_CUDA} for different image and deformation grid resolutions. Realistic speedups are in the range between $1.13$ and $10.1$. Relative GPU performance benefits from higher resolutions, with a maximum speedup at the full image size of $4096^2$ pixels and a deformation grid size of $2049^2$.

\paragraph{Full registration}

We performed a full multi-level registration with three levels for different image and deformation grid sizes analogous to Section~\ref{sec:full_registration}. The input consisted of patches of $4096^2$ pixels in size from histological serial sections of tumor tissue, which were differently stained  and scanned in a high resolution \cite{lotz2016patch}, as shown in Figure~\ref{fig:histo}. Here, image registration allows to align adjacent slices, evaluate different stains on the same slice and to compensate for deformation from the slicing process.
Before registration, the images were rigidly pre-aligned and converted to gray scale.

The matrix-free GPU implementation is compared to the corresponding matrix-free CPU implementation in Figure~\ref{table:multiLevelRuntime_CUDA}. We used a maximum of $20$ iterations per level to ensure comparable runtimes. Depending on the image size, speedups from $1.04$ to $9.21$ are achieved. As before, compared to the CPU code, the highest speedup of $9.21$ is achieved at the finest image resolution of $4096^2$ pixels, since more parallel threads can be distributed on the GPU CUDA cores, resulting in a better GPU utilization.

\paragraph{Summary} 

While the proposed approach is well-suited for GPUs due to its high parallelizability, implementation requires careful consideration of the architecture's specifics. In particular, available memory types need to be fully utilized and unnecessary transfers avoided. While the slightly reduced accuracy has to be accounted for, other features such as atomics can be used to reduce the code complexity without sacrificing performance. Overall, we achieved realistic speedups of about one order of magnitude compared to a CPU implementation.

\subsection{Medical applications} \label{sec:medical_applications}
For the CPU implementation of the matrix-free algorithm, we additionally studied two medical applications of the proposed approach in order to demonstrate the suitability for real-world tasks and to show its potential for clinical use. We used the L-BFGS method for these experiments -- while in practice it required more iterations than the Gauss-Newton scheme, overall runtime was faster.
\subsubsection{Pulmonary image registration} \label{sec:pulmonary_image_registration}
This medical application requires registration of maximum inhale and exhale images of four-dimensional thorax CT (4DCT) scans. The results can be used to assess local lung ventilation and aid the planning of radiation therapy of lung cancer~\cite{kabus2008lung}.

For benchmarking, we used the publicly available DIR-Lab database~\cite{castillo2009framework, castillo2010four}, which consists of ten 4DCT scans. The maximum inhale and exhale images come with 300 expert-annotated landmarks each for evaluation of the registration accuracy. All images have voxel sizes of approximately $\SI{1}{\milli\meter}\times \SI{1}{\milli\meter}\times \SI{2.5}{\milli\meter}$.

As is common in pulmonary image registration, prior to registration the lungs were segmented and the images were cropped to the lung region~\cite{murphy2011evaluation}. The segmentation masks were generated with the fully automatic algorithm proposed in~\cite{lassen2011lung}. 
The cropped images were first coarsely registered with an affine-linear transformation model, the final result serving as an initial guess for the main deformable registration. Prior to deformable registration, the images were isotropically resampled in $z$ direction to match the $x$-$y$ plane resolution.

A multi-resolution scheme with four levels was used, with the finest level being the image at full resolution. The deformation resolution was chosen four times coarser than the image resolution on the finest level. For each coarser level, the number of cells per dimension was halved. The regularizer weight was set to  $\alpha=1$ and the  NGF filter parameters to $\varrho,\tau=10$. The parameters were determined manually by evaluating several different parameter sets, and we also found them to be suitable for lung registration in other applications.

Table~\ref{table:dirlab} summarizes the results and measured runtimes using the L-BFGS optimization method. At an average runtime of \SI{9.23}{\second}, the average landmark error is below the voxel diameter at \SI{0.93}{\milli\meter}. The average landmark error is also the lowest reported in the public benchmark \cite{dirlab} at the time of writing. In comparison, the most recent submission to the benchmark in \cite{vishnevskiy2017isotropic} reports an average landmark error of $0.95 \pm 1.15 \si{\milli\meter}$ with a much longer average runtime of \SI{180}{\second}. Two other methods \cite{polzin2013combining, hermann2014evaluation} achieve errors of $0.95 \pm 1.07 \si{\milli\meter}$. Only \cite{hermann2014evaluation} reports a runtime, stating \SI{98}{\second} on average. 

\begin{table*}
\centering
\begin{tabular}{lS[table-format=2.2,table-figures-uncertainty=1]S[table-format=2.2,table-figures-uncertainty=1]S[table-format=2.2]}
\multicolumn{1}{c}{Dataset}&{Initial (\si{\milli\meter})}&{Proposed (\si{\milli\meter})}&{Runtime (\si{\second})}\\
\toprule
4DCT1 &3.89 \pm 2.78&0.81 \pm 0.89& 6.68\\
4DCT2 &4.34 \pm 3.90&0.75 \pm 0.85& 8.56\\
4DCT3 &6.94 \pm 4.05&0.92 \pm 1.05& 6.78\\
4DCT4 &9.83 \pm 4.86&1.34 \pm 1.28& 7.08\\
4DCT5 &7.48 \pm 5.51&1.06 \pm 1.44& 7.74\\
\midrule
4DCT6 &10.89 \pm 6.97&0.86 \pm 0.96&11.64\\
4DCT7 &11.03 \pm 7.43&0.83 \pm 1.01&12.30\\
4DCT8 &14.99 \pm 9.01&1.02 \pm 1.32&13.75\\
4DCT9 &7.92 \pm 3.98 &0.88 \pm 0.94& 7.85\\
4DCT10&7.30 \pm 6.35 &0.84 \pm 1.00& 9.93\\
\midrule                                                     
\multicolumn{1}{c}{Mean}&8.46 \pm 5.48&\bfseries 0.93 \pm 1.07&\bfseries 9.23\\
 \end{tabular}
\caption{Average landmark errors in millimeters and execution times on CPU in seconds for the DIR-Lab 4DCT datasets after affine-linear pre-alignment (Initial) and after deformable registration (Proposed). Compared to other results reported on the benchmark page \cite{dirlab}, the proposed algorithm achieves the lowest mean distance at the time of writing, with an average runtime of \SI{9.23}{\second}, see also Section~\ref{sec:pulmonary_image_registration} for a comparison.
}
\label{table:dirlab}
\end{table*}

\subsubsection{Oncological follow-up} \label{sec:oncological_followup}
As a second medical application of the proposed registration scheme, we considered oncological follow-up in the  thorax-abdomen region. Here, CT scans are acquired at different time points -- typically a few months apart -- with the goal of assessing the development of tumors, e.g., during chemotherapy. Deformable registration is then employed to propagate prior findings to the current scan, facilitate side-by-side comparison of the same structures at multiple time points via cursor or slice synchronization, and to visualize and highlight change by image subtraction.

On state-of-the-art scanners, the resolution of such CT scans is approximately \SI{0.7}{\milli\meter} isotropic, leading to a challenging image size of $\approx 512\times512\times 900$, depending on the size of the scanned region.
To the best of our knowledge, no public data with expert annotations is available for evaluating registration accuracy. We therefore restricted ourselves to visual assessment and runtime comparison on images from clinical routine (first scan: $512\times 512\times 848$ voxels, second scan: $512\times 512\times 833$ voxels), acquired approximately nine months apart.

A multi-resolution scheme with three levels was used, with one quarter of the original image resolution at the finest level. On each level, the deformation resolution was half the current image resolution. Further increasing the deformation resolution did not improve results, which is also supported by the findings in~\cite{kohn2006gpu,polzin2016memory}. The model parameters were manually chosen as $\alpha=10$ and $\varrho,\tau=5$.

Using the proposed method, we obtain qualitatively satisfying results (Figure~\ref{fig:thoraxReg}) in a clinically acceptable runtime of \SI{12.6}{\second}, enabling online registration of thorax-abdomen CT scans for follow-up examinations. 

\section{Summary and conclusions}\label{sec:discussion}

We presented a novel computational approach for deformable image registration based on the widely-used
normalized gradient fields distance measure and curvature regularization. Through detailed mathematical analysis of the building blocks, we derived a new matrix-free formulation, which tackles two main bottlenecks of the employed registration algorithm:

Firstly, it enables \emph{effective parallelization}. The most-used components exhibit virtually linear scalability (Section \ref{sec:ex_scalability}), efficiently utilizing multiple computational cores on modern CPUs. This leads to an average reduction in overall runtime by a factor of $45.1$ for L-BFGS optimization (Gauss-Newton: $17.3$) compared to a MATLAB implementation, and $9.19$ for L-BFGS optimization (Gauss-Newton: $4.69$) compared to a sparse matrix-based, optimized C++ implementation, despite necessary re-computations of intermediate values. Our GPU-based implementation of the matrix-free registration approach, exploiting the massively parallel architecture, achieves a further speedup of $9.21$ compared to an optimized CPU-based implementation.

Secondly, the presented method largely reduces the \emph{memory requirements} of the registration algorithm. 
As shown in Table~\ref{table:objfunGradRuntime}, even medium resolutions cannot be handled using matrix-based approaches, while the matrix-free scheme is ultimately only limited by runtime. 

Experiments showed that the presented algorithm is able to solve real-world clinical registration tasks with high accuracy at fast runtimes. In the DIR-Lab 4DCT benchmark for lung registration, we achieve the best mean distance among all submissions at an average runtime of \SI{9.23}{\second}. In thorax-abdomen registration for oncology screening, large images could be processed within \SI{12.6}{\second}, potentially allowing for online assessment by clinicians.

The presented scheme allows for efficient image registration on much higher resolutions than previously possible and shows a substantial speedup over existing implementations. It is not tied to a specific platform or CPU type and very suitable for implementation on GPUs, enabling a wider use of rapid deformable image registration in various applications.

\section*{Acknowledgments}
We thank Martin Meike for providing the GPU implementation developed in his thesis \cite{meike2016gpu}. 
\bibliographystyle{siamplain}
\bibliography{literature}

\begin{thebibliography}{10}

\bibitem{avants2008symmetric}
{\sc B.~B. Avants, C.~L. Epstein, M.~Grossman, and J.~C. Gee}, {\em {S}ymmetric
  diffeomorphic image registration with cross-correlation: evaluating automated
  labeling of elderly and neurodegenerative brain}, Med. Image. Anal., 12
  (2008), pp.~26--41.

\bibitem{berg2014highly}
{\sc R.~Berg, L.~K\"{o}nig, J.~R\"{u}haak, R.~Lausen, and B.~Fischer}, {\em
  Highly efficient image registration for embedded systems using a distributed
  multicore {DSP} architecture}, J. Real-Time Image Proc.,  (2014), pp.~1--21.

\bibitem{broit1981optimal}
{\sc C.~Broit}, {\em Optimal Registration of Deformed Images}, PhD thesis,
  1981.

\bibitem{bui2009performance}
{\sc P.~Bui and J.~Brockman}, {\em Performance analysis of accelerated image
  registration using {GPGPU}}, in {Proceedings of 2nd Workshop on General
  Purpose Processing on Graphics Processing Units}, ACM, 2009, pp.~38--45.

\bibitem{burger2013hyperelastic}
{\sc M.~Burger, J.~Modersitzki, and L.~Ruthotto}, {\em {A} hyperelastic
  regularization energy for image registration}, SIAM J. Sci. Comput., 35
  (2013), pp.~B132--B148.

\bibitem{castillo2010four}
{\sc E.~Castillo, R.~Castillo, J.~Martinez, M.~Shenoy, and T.~Guerrero}, {\em
  {F}our-dimensional deformable image reg. using trajectory modeling}, Phys.
  Med. Biol., 55 (2010), p.~305.

\bibitem{dirlab}
{\sc R.~Castillo}, {\em {DIR-Lab Results}}.
\newblock \url{https://www.dir-lab.com/Results.html}.
\newblock Accessed: 2017-04-10.

\bibitem{castillo2009framework}
{\sc R.~Castillo, E.~Castillo, R.~Guerra, V.~E. Johnson, T.~McPhail, A.~K.
  Garg, and T.~Guerrero}, {\em {A} framework for evaluation of deformable image
  registration spatial accuracy using large landmark point sets}, Phys. Med.
  Biol., 54 (2009), p.~1849.

\bibitem{castro2003fair}
{\sc C.~R. Castro-Pareja, J.~M. Jagadeesh, and R.~Shekhar}, {\em {FAIR}: {A}
  hardware architecture for real-time {3-D} image registration}, IEEE T. Inf.
  Technol. B., 7 (2003), pp.~426--434.

\bibitem{collignon1995automated}
{\sc A.~Collignon, F.~Maes, D.~Delaere, D.~Vandermeulen, P.~Suetens, and
  G.~Marchal}, {\em {A}utomated multi-modality image registration based on
  information theory}, in {I}nformation {P}rocessing in {M}edical {I}maging,
  vol.~3, 1995, pp.~264--274.

\bibitem{deluca2015liver}
{\sc V.~De~Luca, T.~Benz, S.~Kondo, L.~K{\"o}nig, D.~L{\"u}bke,
  S.~Rothl{\"u}bbers, O.~Somphone, S.~Allaire, M.~L. Bell, D.~Chung, et~al.},
  {\em The 2014 liver ultrasound tracking benchmark}, Phys. Med. Biol., 60
  (2015), p.~5571.

\bibitem{eklund2013medical}
{\sc A.~Eklund, P.~Dufort, D.~Forsberg, and S.~M. LaConte}, {\em {M}edical
  image processing on the {GPU} -- {P}ast, present and future}, Med. Image.
  Anal., 17 (2013), pp.~1073 -- 1094.

\bibitem{fischer2002fast}
{\sc B.~Fischer and J.~Modersitzki}, {\em {F}ast diffusion registration},
  Contemp. Math., 313 (2002), pp.~117--128.

\bibitem{fischer2003curvature}
{\sc B.~Fischer and J.~Modersitzki}, {\em {C}urvature based image
  registration}, J. Math. Imaging Vision, 18 (2003), pp.~81--85.

\bibitem{galban2012computed}
{\sc C.~J. Galb{\'a}n, M.~K. Han, J.~L. Boes, K.~A. Chughtai, C.~R. Meyer,
  T.~D. Johnson, S.~Galb{\'a}n, A.~Rehemtulla, E.~A. Kazerooni, F.~J. Martinez,
  et~al.}, {\em Computed tomography-based biomarker provides unique signature
  for diagnosis of {COPD} phenotypes and disease progression}, Nat. Med., 18
  (2012), pp.~1711--1715.

\bibitem{gigengack2012motion}
{\sc F.~Gigengack, L.~Ruthotto, M.~Burger, C.~H. Wolters, X.~Jiang, and K.~P.
  Schafers}, {\em {M}otion correction in dual gated cardiac {PET} using
  mass-preserving image registration}, IEEE T. Med. Imaging, 31 (2012),
  pp.~698--712.

\bibitem{haber2007octree}
{\sc E.~Haber, S.~Heldmann, and J.~Modersitzki}, {\em {A}n octree method for
  parametric image registration}, SIAM J. Sci. Comput., 29 (2007),
  pp.~2008--2023.

\bibitem{haber2008adaptive}
{\sc E.~Haber, S.~Heldmann, and J.~Modersitzki}, {\em {A}daptive mesh
  refinement for nonparametric image registration}, SIAM J. Sci. Comput., 30
  (2008), pp.~3012--3027.

\bibitem{haber2005beyond}
{\sc E.~Haber and J.~Modersitzki}, {\em Beyond mutual information: A simple and
  robust alternative}, in Bildverarbeitung f{\"u}r die Medizin, 2005,
  pp.~350--354.

\bibitem{harris2007optimizing}
{\sc M.~Harris}, {\em {Optimizing parallel reduction in CUDA}}, NVIDIA
  Developer Technology,  (2007).

\bibitem{heldmann2006non}
{\sc S.~Heldmann}, {\em Non-linear Registration Based on Mutual Information:
  Theory, Numerics, and Application}, Logos-Verlag, Berlin, 2006.

\bibitem{hermann2014evaluation}
{\sc S.~Hermann}, {\em {Evaluation of scan-line optimization for 3D medical
  image registration}}, 2014, pp.~3073--3080.

\bibitem{kabus2008lung}
{\sc S.~Kabus, J.~von Berg, T.~Yamamoto, R.~Opfer, and P.~J. Keall}, {\em Lung
  ventilation estimation based on 4{D}-{CT} imaging}, in First International
  Workshop on Pulmonary Image Analysis, 2008, pp.~73--81.

\bibitem{kohn2006gpu}
{\sc A.~K{\"o}hn, J.~Drexl, F.~Ritter, M.~K{\"o}nig, and H.-O. Peitgen}, {\em
  {GPU} accelerated image registration in two and three dimensions}, in BVM,
  2006, pp.~261--265.

\bibitem{konig2015parallel}
{\sc L.~K\"{o}nig, A.~Derksen, M.~Hallmann, and N.~Papenberg}, {\em Parallel
  and memory efficient multimodal image registration for radiotherapy using
  normalized gradient fields}, in IEEE International Symposium on Biomedical
  Imaging: From Nano to Macro, 2015, pp.~734--738.

\bibitem{konig2016deformable}
{\sc L.~K{\"o}nig, A.~Derksen, N.~Papenberg, and B.~Haas}, {\em Deformable
  image registration for adaptive radiotherapy with guaranteed local rigidity
  constraints}, Radiat. Oncol., 11 (2016), p.~122.

\bibitem{konig2014nonlinear}
{\sc L.~K{\"o}nig, T.~Kipshagen, and J.~R{\"u}haak}, {\em A non-linear image
  registration scheme for real-time liver ultrasound tracking using normalized
  gradient fields}, in Proc. MICCAI CLUST14, Boston, USA, 2014, pp.~29--36.

\bibitem{konig2014fast}
{\sc L.~K\"{o}nig and J.~R\"{u}haak}, {\em A fast and accurate parallel
  algorithm for non-linear image registration using normalized gradient
  fields}, in IEEE International Symposium on Biomedical Imaging: From Nano to
  Macro, 2014, pp.~580--583.

\bibitem{lassen2011lung}
{\sc B.~Lassen, J.-M. Kuhnigk, M.~Schmidt, S.~Krass, and H.-O. Peitgen}, {\em
  {L}ung and lung lobe segmentation methods at {F}raunhofer {MEVIS}}, in Fourth
  International Workshop on Pulmonary Image Analysis, 2011, pp.~185--200.

\bibitem{lemoigne2011image}
{\sc J.~Le~Moigne, N.~S. Netanyahu, and R.~D. Eastman}, {\em Image Registration
  for Remote Sensing}, Cambridge University Press, Cambridge, 2011.

\bibitem{lotz2016patch}
{\sc J.~Lotz, J.~Olesch, B.~Muller, T.~Polzin, P.~Galuschka, J.~M. Lotz,
  S.~Heldmann, H.~Laue, M.~Gonzalez-Vallinas, A.~Warth, et~al.}, {\em
  {Patch-based nonlinear image registration for gigapixel whole slide images}},
  IEEE T. Bio-Med. Eng, 63 (2016), pp.~1812--1819.

\bibitem{mang2017semi}
{\sc A.~Mang and G.~Biros}, {\em {A Semi-Lagrangian two-level preconditioned
  Newton-Krylov solver for constrained diffeomorphic image registration}}, SIAM
  J. Sci. Comput., 39 (2017), pp.~B1064--B1101.

\bibitem{mang2016distributed}
{\sc A.~Mang, A.~Gholami, and G.~Biros}, {\em {Distributed-memory large
  deformation diffeomorphic 3D image registration}}, 2016, pp.~842--853.

\bibitem{meike2016gpu}
{\sc M.~Meike}, {\em {GPU-basierte nichtlineare Bildregistrierung}}, {Master's
  thesis}, University of L{\"{u}}beck, 2016.

\bibitem{modersitzki2004}
{\sc J.~Modersitzki}, {\em {N}umerical {M}ethods for {I}mage {R}egistration},
  Oxford University Press, Oxford, 2004.

\bibitem{FAIR}
{\sc J.~Modersitzki}, {\em {FAIR}: {F}lexible {A}lgorithms for {I}mage
  {R}egistration}, vol.~6, SIAM, Philadelphia, PA, 2009.

\bibitem{murphy2011evaluation}
{\sc K.~Murphy, B.~Van~Ginneken, J.~M. Reinhardt, S.~Kabus, K.~Ding, X.~Deng,
  K.~Cao, K.~Du, G.~E. Christensen, V.~Garcia, et~al.}, {\em {E}valuation of
  registration methods on thoracic {CT}: the {EMPIRE}10 challenge}, IEEE T.
  Med. Imaging, 30 (2011), pp.~1901--1920.

\bibitem{muyan2008fast}
{\sc P.~Muyan-Ozcelik, J.~D. Owens, J.~Xia, and S.~S. Samant}, {\em Fast
  deformable registration on the {GPU}: {A} {CUDA} implementation of demons},
  in {International Conference on Computational Sciences and Its Applications
  (ICCSA), 2008}, IEEE, 2008, pp.~223--233.

\bibitem{nocedal1999numerical}
{\sc J.~Nocedal and S.~Wright}, {\em Numerical Optimization}, Springer, New
  York, NY, 1999.

\bibitem{nvidia2017cuda}
{\sc {NVIDIA Corporation}}, {\em {CUDA C Programming Guide}},
  no.~PG-02829-001{\_}v8.0, 2017,
  \url{http://docs.nvidia.com/cuda/pdf/CUDA\_C\_Programming\_Guide.pdf}.

\bibitem{polzin2016memory}
{\sc T.~Polzin, M.~Niethammer, M.~P. Heinrich, H.~Handels, and J.~Modersitzki},
  {\em Memory efficient {LDDMM} for lung {CT}}, in Medical Image Computing and
  Computer-Assisted Intervention -- MICCAI 2016, 2016, pp.~28--36.

\bibitem{polzin2013combining}
{\sc T.~Polzin, J.~R{\"{u}}haak, R.~Werner, J.~Strehlow, S.~Heldmann,
  H.~Handels, and J.~Modersitzki}, {\em {Combining automatic landmark detection
  and variational methods for lung CT registration}}, 2013, pp.~85--96.

\bibitem{ruhaak2013highly}
{\sc J.~R{\"u}haak, S.~Heldmann, T.~Kipshagen, and B.~Fischer}, {\em Highly
  accurate fast lung {CT} registration}, in SPIE Medical Imaging, Image
  Processing, International Society for Optics and Photonics, 2013,
  pp.~86690Y--86690Y--9.

\bibitem{ruhaak2017matrix}
{\sc J.~R{\"u}haak, L.~K{\"o}nig, F.~Tramnitzke, H.~K{\"o}stler, and
  J.~Modersitzki}, {\em A matrix-free approach to efficient affine-linear image
  registration on {CPU} and {GPU}}, J. Real-Time Image Proc., 13 (2017),
  pp.~205--225.

\bibitem{shackleford2013high}
{\sc J.~Shackleford, N.~Kandasamy, and G.~Sharp}, {\em High Performance
  Deformable Image Registration Algorithms for Manycore Processors}, Morgan
  Kaufmann, Waltham, MA, 2013.

\bibitem{shamonin2013fast}
{\sc D.~P. Shamonin, E.~E. Bron, B.~P. Lelieveldt, M.~Smits, S.~Klein, and
  M.~Staring}, {\em {F}ast parallel image registration on {CPU} and {GPU} for
  diagnostic classification of {A}lzheimer's disease}, Front. Neuroinform., 7
  (2013), p.~50.

\bibitem{shams2010survey}
{\sc R.~Shams, P.~Sadeghi, R.~Kennedy, and R.~Hartley}, {\em {A} survey of
  medical image registration on multicore and the {GPU}}, IEEE Signal Proc.
  Mag., 27 (2010), pp.~50--60.

\bibitem{sturmer2008fast}
{\sc M.~St{\"u}rmer, H.~K{\"o}stler, and U.~R{\"u}de}, {\em A fast full
  multigrid solver for applications in image processing}, Numer. Linear Algebra
  Appl., 15 (2008), pp.~187--200.

\bibitem{stutzer2016evaluation}
{\sc K.~St{\"u}tzer, R.~Haase, F.~Lohaus, S.~Barczyk, F.~Exner, S.~L{\"o}ck,
  J.~R{\"u}haak, et~al.}, {\em Evaluation of a deformable registration
  algorithm for subsequent lung computed tomography imaging during
  radiochemotherapy}, Med. Phys., 43 (2016), pp.~5028--5039.

\bibitem{vercauteren2009diffeomorphic}
{\sc T.~Vercauteren, X.~Pennec, A.~Perchant, and N.~Ayache}, {\em
  {D}iffeomorphic demons: {E}fficient non-parametric image registration},
  NeuroImage, 45 (2009), pp.~S61--S72.

\bibitem{viola1997alignment}
{\sc P.~Viola and W.~Wells~III}, {\em {A}lignment by maximization of mutual
  information}, Int. J. Comput. Vis., 24 (1997), pp.~137--154.

\bibitem{vishnevskiy2017isotropic}
{\sc V.~Vishnevskiy, T.~Gass, G.~Szekely, C.~Tanner, and O.~Goksel}, {\em
  {Isotropic total variation regularization of displacements in parametric
  image registration}}, IEEE T. Med. Imaging, 36 (2017), pp.~385--395.

\end{thebibliography}
\end{document}